\documentclass{article}

\usepackage{wrapfig}

\usepackage[preprint]{neurips_2025}

\usepackage[utf8]{inputenc} 
\usepackage[T1]{fontenc}    
\usepackage{hyperref}       
\usepackage{url}            
\usepackage{booktabs}       
\usepackage{amsfonts}       
\usepackage{nicefrac}       
\usepackage{microtype}      
\usepackage{xcolor}         
\usepackage{siunitx}
\usepackage{float}
\usepackage{graphicx}
\usepackage{caption}        
\usepackage{subcaption}     
\usepackage{multirow}
\usepackage{tabularx} 
\newcommand{\ie}{\textit{i.e.}}

\usepackage{amsmath} 
\usepackage{tcolorbox}

\tcbset{
  mybox/.style={
    colback=white,
    colframe=black,
    boxrule=0.6pt,
    arc=1mm,
    leftrule=0.6pt,
    rightrule=0.6pt,
    toptitle=1mm,
    bottomtitle=1mm,
    fonttitle=\bfseries,
    coltitle=black
  },
  torchbox/.style={
    colback=white,
    colframe=gray!70,
    boxrule=0.6pt,
    arc=1mm,
    leftrule=0.6pt,
    rightrule=0.6pt,
    toptitle=1mm,
    bottomtitle=1mm,
    fonttitle=\bfseries,
    coltitle=black
  }
}

\usepackage{listings}          
\usepackage{enumitem}

\title{Towards Omnidirectional Reasoning with 360-R1: \\ A Dataset, Benchmark, and GRPO-based Method}

\author{
Xinshen Zhang$^{1}$ \quad Zhen Ye$^{2}$ \quad Xu Zheng$^{3}$\thanks{Corresponding author: Xu Zheng \texttt{<zhengxu128@gmail.com>}} \\
\\
$^{1}$The Hong Kong Polytechnic University \quad
$^{2}$HKUST \quad
$^{3}$HKUST(GZ)
}

\begin{document}
\maketitle

\begin{figure}[H]
  \begin{center}
    \includegraphics[width=1\textwidth]{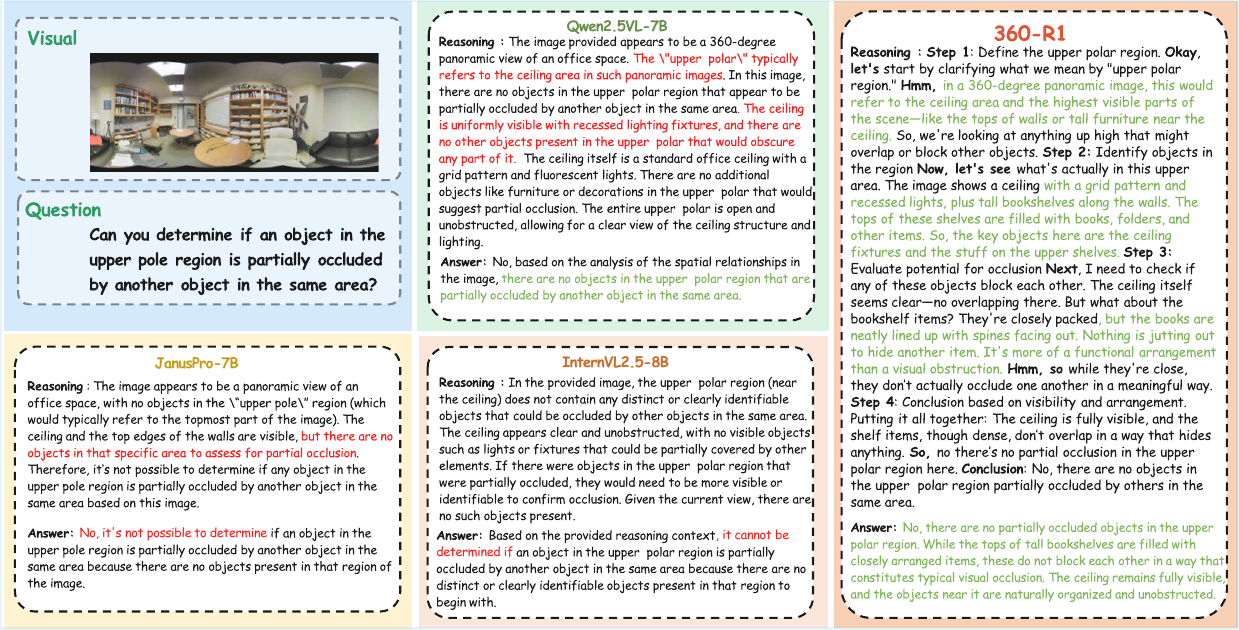}
  \end{center}
  \caption{
\textbf{Comparison of occlusion reasoning in 360° VQA.} Given a panoramic scene and a spatial question about whether an object in the upper pole region is partially occluded, four models provide different levels of reasoning. \textbf{360-R1} demonstrates the most precise and comprehensive reasoning, identifying relevant spatial elements and producing a correct answer. \textbf{QwenVL2.5-7B} gives the correct answer but its explanation is partially flawed and lacks depth. In contrast, both \textbf{JanusPro-7B} and \textbf{InternVL2.5-8B} fail to answer correctly, primarily due to limited or inaccurate analysis of the upper pole region. 
}
  \vspace{-10pt}
  \label{fig:CaseFigure}
\end{figure}

\begin{abstract}
Omnidirectional images (ODIs), with their 360° field of view, provide unparalleled spatial awareness for immersive applications like augmented reality and embodied AI. However, the capability of existing multi-modal large language models (MLLMs) to comprehend and reason about such panoramic scenes remains underexplored. This paper addresses this gap by introducing \textbf{\textit{OmniVQA}}, the \textbf{\textit{first}} dataset and conducting the \textbf{\textit{first}} benchmark for omnidirectional visual question answering. Our evaluation of state-of-the-art MLLMs reveals significant limitations in handling omnidirectional visual question answering, highlighting persistent challenges in object localization, feature extraction, and hallucination suppression within panoramic contexts. These results underscore the disconnect between current MLLM capabilities and the demands of omnidirectional visual understanding, which calls for dedicated architectural or training innovations tailored to 360 ° imagery. Building on the OmniVQA dataset and benchmark, we further introduce a rule-based reinforcement learning method, \textbf{\textit{360-R1}}, based on Qwen2.5-VL-Instruct. Concretely, we modify the group relative policy optimization (GRPO) by proposing three novel reward function, \ie, reasoning process similarity reward, answer semantic accuracy reward, and structured format compliance reward. Extensive experiments on our OmniVQA demonstrate the superiority of our proposed method in omnidirectional space (+6\%$\uparrow$). 
\end{abstract}

\begin{figure}[h!]
  \begin{center}
    \includegraphics[width=1\textwidth]{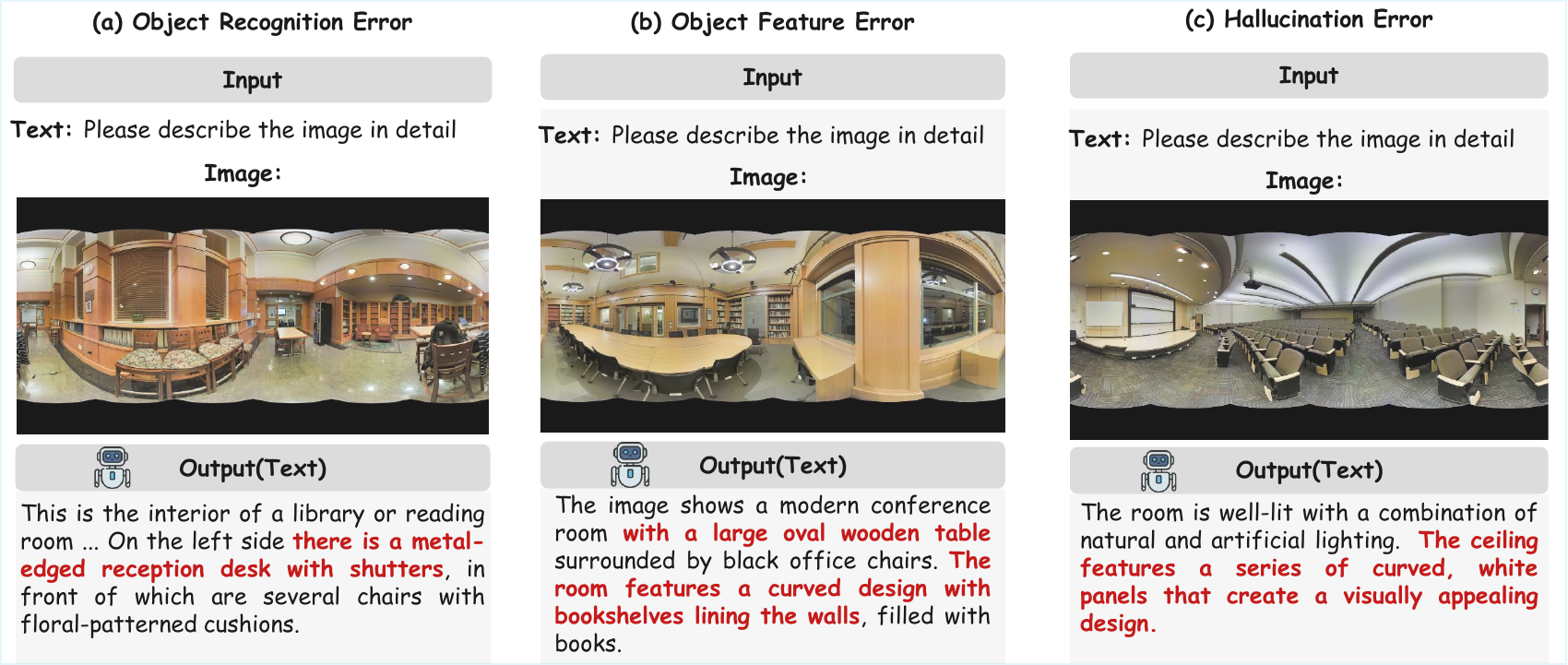}
  \end{center}
  \caption{
    \textbf{Error Cases in Omnidirectional Captioning.} Three common errors by multimodal LLMs on 360° images: (a) misidentified objects; (b) incorrect object attributes or context; (c) hallucinated content unrelated to the image.
  }
  \vspace{-16pt}
  \label{fig:Figure1}
\end{figure}
\section{Introduction}
Driven by the rapid advancement of Augmented Reality (AR), Virtual Reality (VR), and Embodied AI systems—as well as the growing demand for immersive visual experiences—omnidirectional images (ODIs) have attracted significant research interest due to their ability to provide a comprehensive perception of surrounding environments~\cite{yan2024survey,ref1,ref2,ref3,ref4,ref5,ref6,Li2022CVPR,shen2022panoformer,shen2021distortion, jiang2021unifuse, DBLP:conf/iclr/WijmansKMLEPSB20, chaplot2020object, ku2020room, krantz_vlnce_2020,zheng2025retrieval,huo2025mmunlearner}. ODIs have been explored across a variety of downstream tasks, such as scene understanding~\cite{DBLP:conf/cvpr/ZhengZLCFW23,DBLP:conf/iccv/ZhengPLW23,DBLP:conf/cvpr/ZhengZVW24,DBLP:journals/pami/ZhengZVW25,DBLP:conf/cvpr/ZhangLZW24,DBLP:journals/corr/abs-2503-07098,cai2024interact360}.
Unlike conventional 2D images with limited fields of view, ODIs capture the complete $180^\circ \times 360^\circ$ surroundings, encoding much richer spatial information, which makes them especially suitable for applications requiring holistic environmental understanding~\cite{ref7,ref8,ref9,Yun2021PanoAVQA,xiao2012sun360}.

Recently, significant progress has been made in developing strong and generalizable Multi-modal Large Language Models (MLLMs)~\cite{ref10,ref11,ref12,ref13,ref14,ref15,ref16,ref18,ref19}. These models are capable of integrating information from diverse modalities—such as text, images, and beyond—and have demonstrated impressive generalization across a wide range of downstream tasks~\cite{ref17,ref20,yue2023mmmu, yue2024mmmu, fang2025creationmmbench,zheng2023deep}. For instance, the Visual Question Answering (VQA) serves as a crucial benchmark for learning usable real-world models and assessing models' ability in integrating visual comprehension of images with semantic understanding of questions, alongside the necessary reasoning capabilities~\cite{DBLP:journals/corr/abs-2305-11033,DBLP:journals/corr/abs-2501-03939}. 

Despite these advances, VQA research remains constrained to conventional images, overlooking the unique advantages of omnidirectional vision, \ie, the comprehensive understanding of the whole surrounding scenes~\cite{VQA,DBLP:journals/corr/abs-1902-09506, schwenk2022aokvqabenchmarkvisualquestion,lu2022learn}. Meanwhile, existing work~\cite{ref21} fails to make their proposed datasets public available and also does not benchmarking the MLLMs' performance on panoramic domain.
In this paper, we propose the first omnidirectional visual question answering dataset, OmniVQA. The dataset focuses on key tasks such as object localization, object attribute analysis, and spatial relationship reasoning  in polar regions. As illustrated in Figure~\ref{fig:Figure1}, persistent challenges remain in accurately performing object recognition, attribute analysis, and spatial relationship description in panoramic views. These limitations lead to subpar performance by current MLLMs, as shown in Table~\ref{tab:metrics}, where JanusPro-7B achieves a relatively low Qwenscore and DeepSeekScore. Furthermore, benchmarks tailored for omnidirectional VQA remain scarce, hindering evaluation and comparison in complex panoramic settings.

To construct the OmniVQA dataset, we leverage equirectangular projection (ERP) panoramic images from the 2D-3D-S dataset~\cite{ref22}. To address the geometric distortions inherent in panoramic imagery, we design three categories of questions: object identification, attribute analysis, and spatial relationship reasoning. Our data generation pipeline consists of three stages. First, a Qwen2.5-VL~\cite{ref23} model fine-tuned on OmniVQABench is used to generate detailed visual descriptions. These descriptions are then passed to DeepSeek-R1~\cite{deepseekr1} to produce Chain-of-Thought (CoT)~\cite{DBLP:journals/corr/abs-2201-11903} style reasoning. Finally, Qwen2.5-14B Instruct~\cite{qwen2025qwen25technicalreport} summarizes the reasoning into a concise answer. To ensure annotation quality, we adopt an iterative refinement strategy. Reasoning–answer pairs from fine-tuned and untuned models are compared using SentenceBERT-Score, and high-quality pairs (score > 0.8) are used to update the model. The process repeats, followed by manual correction of the remaining samples.

We also propose a post-training strategy based on rule-based reinforcement learning, termed 360-R1. Built upon the Qwen2.5-VL-Instruct model~\cite{ref23}, which has already undergone extensive instruction tuning,360-R1 introduces structured reward functions specifically tailored for 360° visual comprehension. These include rewards for reasoning consistency, semantic answer accuracy, and output format compliance, all the semantic similarity automatically evaluated via the DeepSeek-V3~\cite{deepseekai2025deepseekv3technicalreport}  . We adopt the Group Relative Policy Optimization (GRPO) ~\cite{shao2024deepseekmathpushinglimitsmathematical}  algorithm to integrate these rewards while ensuring training stability. This strategy significantly enhances the model’s ability to perform spatial reasoning, suppress hallucinations, and generate machine-readable responses in complex panoramic scenes~\ref{fig:CaseFigure}.
\textbf{(I) \textit{OmniVQA Dataset}.} We introduce the first open-source omnidirectional VQA dataset built upon ERP-format panoramic images, containing three task types: object identification, attribute analysis, and spatial reasoning—particularly focused on polar regions. 
\textbf{(II) \textit{OmniVQABench}.} A comprehensive benchmark designed to evaluate MLLMs in 360° visual environments.
\textbf{(III) \textit{360-R1 Post-training Framework}.} A rule-guided reinforcement learning framework leveraging structured rewards and GRPO, significantly enhancing spatial reasoning and  object identification.

\section{Related Work}
\noindent \textbf{Omnidirectional Vision Question Answering (OVQA)} 
Visual Question Answering (VQA) aims to generate natural language answers given an image and a corresponding question~\cite{DBLP:journals/corr/abs-2305-11033,DBLP:journals/corr/abs-2501-03939}. The widely used VQA v1.0~\cite{VQA} includes both real-world and abstract scenes. VQA v2.0~\cite{Goyal_2017_CVPR} mitigates language bias by balancing question distributions per image. To extend VQA to omnidirectional vision, the VQA 360° dataset~\cite{ref21} adapts Stanford 2D-3D~\cite{ref22} and Matterport3D~\cite{Matterport3D}, using cube projection (CP) for panoramic representation. However, the dataset is not publicly released. In contrast, Pano-AVQA~\cite{Yun2021PanoAVQA} targets 360° video-based QA, incorporating bounding-box-level annotations and enabling spherical spatial and audiovisual reasoning. Despite these efforts, publicly available, high-quality benchmarks for image-based omnidirectional VQA remain scarce, limiting thorough evaluation of models in panoramic scenarios. To address this gap, we introduce \textbf{OmniVQA}, the first open-source omnidirectional VQA dataset and benchmark, focusing on object identification, attribute analysis, and spatial reasoning in complex 360° environments.

\noindent \textbf{Multi-Modal Large Language Models}
Multi-modal large language models (MLLMs) have shown strong performance in visual dialogue, image captioning, and visual question answering~\cite{lyu2024unibind,DBLP:journals/corr/abs-2201-12086,li2023blip2bootstrappinglanguageimagepretraining,caffagni2024revolutionmultimodallargelanguage,DBLP:journals/corr/DasKGSYMPB16,DBLP:journals/corr/XuBKCCSZB15,lyu2024omnibind}.
InternVL 2.5 retains the ViT-MLP-LLM architecture while reducing visual tokens via pixel rearrangement, supporting multi-image and video input with techniques like dynamic high-resolution training and loss re-weighting~\cite{Internvl2_5}.
LLaVA-OneVision~\cite{llava-onevision}, the first open-source model effective across single-image, multi-image, and video tasks, combines Qwen-2~\cite{qwen2} and SigLIP~\cite{zhai2023sigmoidlosslanguageimage} using MLP, with a Higher AnyRes strategy for balanced performance and efficiency.
Qwen2.5VL~\cite{ref23} introduces 2D-RoPE, window attention, dynamic FPS sampling, and MLP-based sequence compression for efficient video understanding.
DeepSeek-VL2~\cite{wu2024deepseekvl2} employs a Mixture-of-Experts architecture with dynamic tiling and latent attention for high-resolution processing.
Janus~\cite{wu2024janus} uses decoupled visual paths within a unified Transformer, while JanusPro~\cite{chen2025januspro} extends this with prolonged pretraining and expanded multi-modal data for improved generation.
In this paper, we propose the \textbf{360-R1} method, which incorporates rule-based reinforcement learning during post-training, specialized for 360° environments.

\noindent \textbf{Reasoning with CoT}
Wei et al.\cite{lu2022learn} pioneered few-shot CoT reasoning with manually annotated rationales. More recently, reasoning-oriented large language models (RLLMs) like OpenAI o1\cite{openai2024o1} and DeepSeek R1~\cite{deepseekr1} have driven extensive research on long CoT reasoning. In the vision-language domain, Vision-R1~\cite{huang2025visionr1} introduces multimodal CoT by converting visual inputs into structured reasoning chains, while R1-OneVision~\cite{yang2025r1onevision} proposes a cross-modal CoT pipeline for step-by-step reasoning. LLaVA-CoT~\cite{xu2024llavacot} uses a multi-stage approach with stage-level beam search for complex multimodal tasks, and LlamaV-o1~\cite{thawakar2025llamavo1} incorporates curriculum learning for efficient multi-step reasoning.
In this paper, our \textbf{OmniVQA} dataset provides two-step CoT annotations—reasoning and answer summaries—to guide the reinforcement learning of the \textbf{360-R1} framework.

\begin{table}[t!]
\centering
\small
\caption{Question types, templates, and examples for the OmniVQA dataset.}
\label{tab:panoramic_qtypes}
\resizebox{\textwidth}{!}{
\begin{tabular}{p{4cm} p{6.5cm} p{5cm}}
\toprule
\textbf{Question Type} & \textbf{Question Template} & \textbf{Question Example} \\
\midrule
\multicolumn{3}{l}{\textbf{(i) Object Identification (1,510 questions, 31.12\%)}} \\
\midrule
\multirow{2}{=}{\parbox{4cm}{Object Identification:\\
Recognizing objects within \\
polar regions of \\
panoramic images.}}
& What object is presented in the [polar region]? 
& What object is presented in the top polar region of the image? \\
\cmidrule(lr){2-3}
& Which objects can be identified in the [polar region]? 
& Which objects can be identified in the bottom polar area? \\
\midrule

\multicolumn{3}{l}{\textbf{(ii) Object Attribute Analysis (1,255 questions, 25.87\%)}} \\
\midrule
\multirow{2}{=}{\parbox{4cm}{Object Attribute Analysis:\\
Describing visual attributes \\
such as characteristics,shape etc.}}
& What shape does the object in the [polar region] exhibit?
& What shape does the object in the top polar region exhibit? \\
\cmidrule(lr){2-3}
& What visual features can you observe about the object in the [polar region]?
& What visual features can you observe about the object in the upper polar area? \\
\midrule

\multicolumn{3}{l}{\textbf{(iii) Spatial Relationship Reasoning (2,087 questions, 43.01\%)}} \\
\midrule
\multirow{2}{=}{\parbox{4cm}{Spatial Relationship\\
Reasoning:\\
Inferring spatial\\
relationships among\\
multiple objects}}\\

& What is the spatial relationship between [object A in the polar region] and [object B near it]?
& What is the spatial relationship between the object in the polar region and the object near it? \\
\cmidrule(lr){2-3}
& What is the spatial relationship between objects in the [polar region]?
& What is the spatial relationship between objects in the top polar regions? \\
\bottomrule
\end{tabular}
}
\vspace{-8pt}
\end{table}

\begin{figure}[ht!]
  \begin{center}
    \includegraphics[width=1\textwidth]{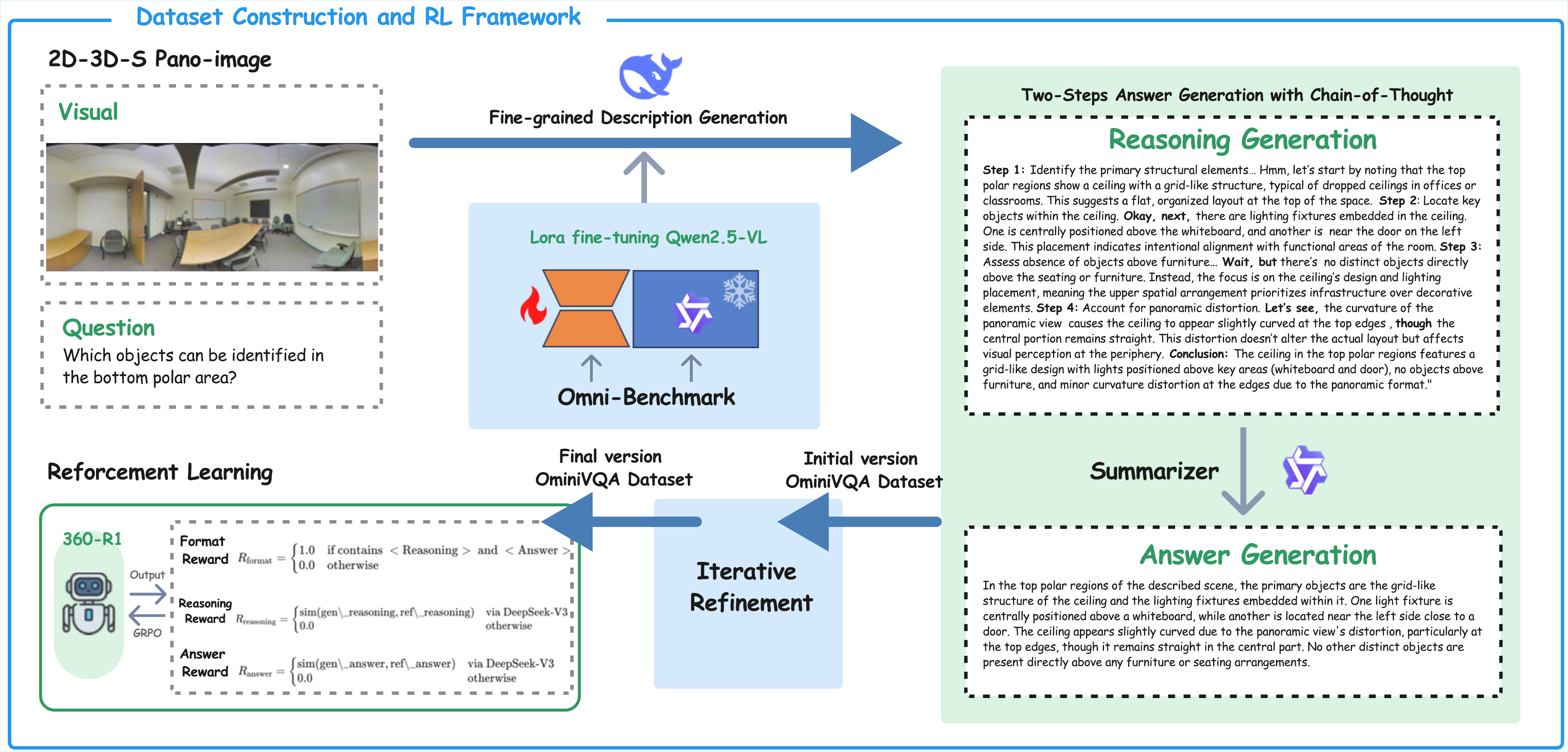}
  \end{center}
  \caption{
    Overview of the OmniVQA Dataset Construction and 360-R1 Framework.
}
\vspace{-12pt}
  \label{fig:framework}
\end{figure}
\section{OmniVQA: Dataset and Benchmark}
\subsection{Overview}
The OmniVQA dataset is the first open-source dataset in the field of omnidirectional visual question answering , introducing novel challenges such as object attribute identification and spatio-temporal reasoning. It extends traditional VQA tasks by incorporating panoramic scene understanding and advanced spatial reasoning, enabling multi-modal interpretation of complex, immersive environments. Developed based on the Stanford 2D-3D-S dataset \cite{ref22}, which covers over 6,000 m² of indoor scenes, OmniVQA dataset contains 1,213 panoramic images in equirectangular projection (ERP) format at a resolution of 1440×720, preserving angular fidelity and geometric structure. A total of 4,852 VQA pairs are included, distributed as follows:31.12\%questions (1,510 ) focus on object identification, 25.87\% (1,255) analyze object attributes, and 43.01\% (2,087) probe spatial relationship reasoning.

\subsection{Dataset Construction}
\noindent \textbf{Question Design.} The OmniVQA dataset is constructed based on  panoramic images from the 2D-3D-S dataset~\cite{ref22}. To address the three typical error types illustrated in Figure~\ref{fig:Figure1}, we design three corresponding types of questions. The first type focuses on object identification, which requires recognizing objects within the panoramic scene, particularly those located in the polar regions where distortions are most prominent. The second type involves object attribute analysis, which asks the model to describe visual characteristics such as shape, size, and material. The third type emphasizes spatial relationship reasoning, where the goal is to infer spatial arrangements, occlusion states, and interactions between objects. Compared with outdoor VQA datasets that typically contain sparse visual content~\cite{streetlearn}, the OmniVQA dataset emphasizes dense indoor scenes. In these environments, the distortions introduced by equirectangular projection in the polar areas significantly hinder accurate visual perception. Our question design specifically addresses these challenges and aims to enhance the model’s ability to understand panoramic scenes under geometric distortions. A detailed breakdown of question types is presented in Table~\ref{tab:panoramic_qtypes}.

\noindent \textbf{Two Steps Answer Generation with Chain-of-Thought~\cite{DBLP:journals/corr/abs-2201-11903}.} To improve the quality of the answer in the presence of distortions in the panoramic image, we adopt a two-step reasoning pipeline, as illustrated in Figure\ref{fig:framework}. In the first step, a Qwen2.5-VL-7B model is fine-tuned using low-rank adaptation (LoRA)~\cite{ref24} on our OmniVQABench dataset to generate detailed visual descriptions of each panoramic image. In the second step, the DeepSeek-R1 model takes these descriptions as input and generates structured CoT reasoning sequences that capture spatial and semantic relationships. Finally, the Qwen2.5-14B model summarizes CoT reasoning into concise answers. This modular pipeline enhances robustness by structuring the reasoning process into distinct perception and reasoning stages, which improves accuracy under panoramic distortion.

\begin{figure}[t!]
  \begin{center}
    \includegraphics[width=1\textwidth]{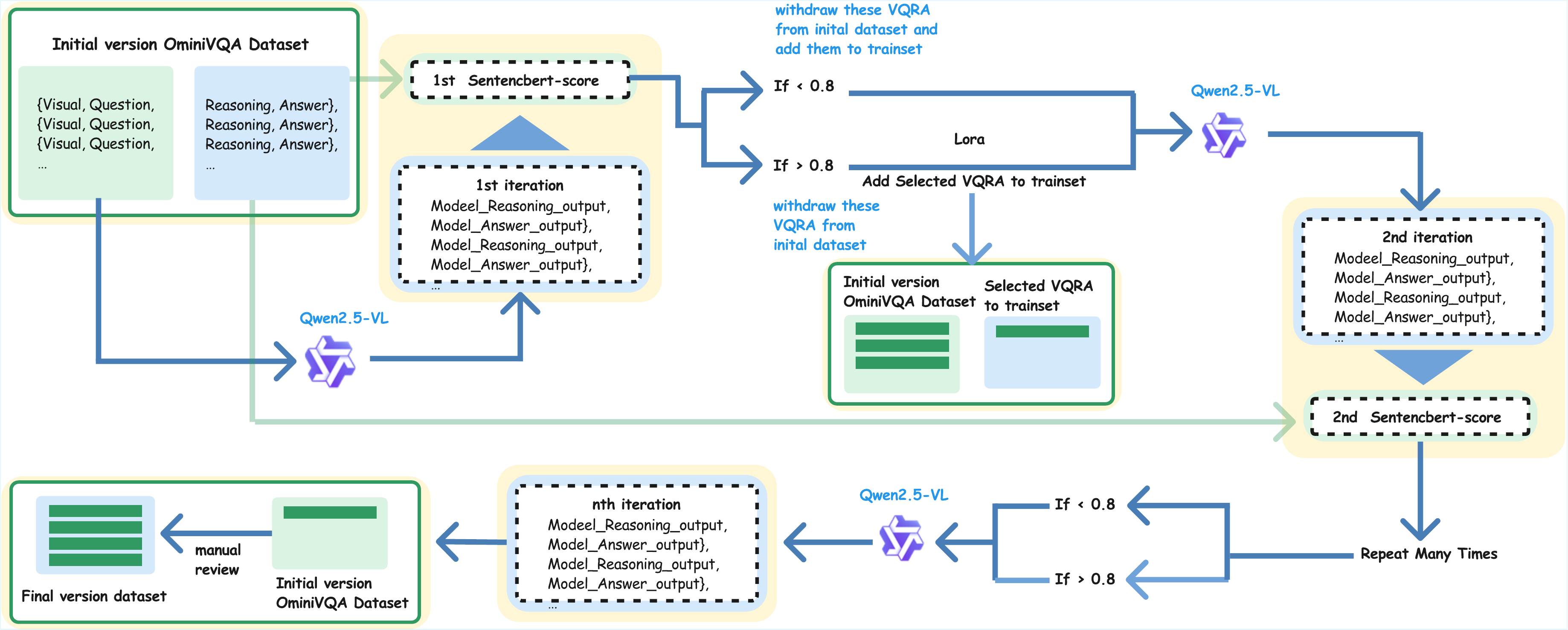}
  \end{center}
  \caption{
    Iterative Refinement Pipeline.
  }
  \label{fig:Iteration}
  \vspace{-16pt}
\end{figure}

\noindent \textbf{Iterative Refinement.} To further improve dataset quality, we we introduce an iterative refinement loop, as shown in Figure~\ref{fig:Iteration}. First, we compare reasoning-answer pairs generated by two models: the fine-tuned Qwen2.5-VL-7B (trained on OmniVQABench) and the original Qwen2.5-VL-7B (without fine-tuning). Both outputs are summarized into final answers using Qwen2.5-14B. Then, we evaluate the semantic similarity between the two outputs using SentenceBERT-Score ~\cite{sentencebert}, applied separately to the reasoning and the answer components. These scores are fused using an F1-style aggregation metric. Pairs with scores above 0.8 are selected to fine-tune a new model instance, initialized from scratch. This updated model is used to generate outputs for the remaining questions, and the evaluation process is repeated. High-confidence pairs are incrementally added to the training set. The cycle continues until only a small set of low-confidence pairs remains, which are manually verified and corrected to ensure overall consistency and accuracy.

\section{Multi-modal Omnidirectional Visual Question Answering Benchmark}

To construct a reliable benchmark for multi-modal omnidirectional VQA, we employ the QwenVL model to generate answers for panoramic images and their corresponding sampled questions. By analyzing these outputs, we identify three major error types, as illustrated in Figure~\ref{fig:Figure1}. Guided by the distribution of these error types, we select relevant images from the Stanford 2D-3D-S dataset~\cite{ref22}, maintaining a 2:1:1 ratio across the three error categories. A few error-free samples are also included to ensure diversity, resulting in a benchmark of 200 panoramic images.
To construct a reliable benchmark for multi-modal omnidirectional VQA, we employ the QwenVL model to generate answers for panoramic images and their corresponding sampled questions. By analyzing these outputs, we identify three major error types, as illustrated in Figure~\ref{fig:Figure1}. Guided by the distribution of these error types, we select relevant images from the Stanford 2D-3D-S dataset~\cite{ref22}, maintaining a 2:1:1 ratio across the three error categories. A few error-free samples are also included to ensure diversity, resulting in a benchmark of 200 panoramic images.
To generate fine-grained visual descriptions for the 200 selected images, GPT-4o~\cite{openai2024gpt4o} is prompted using several high-quality  examples.  All descriptions are manually verified, with particular focus on correcting panoramic distortions and restoring object-level details. The DeepSeek-R1 model takes these descriptions as input and generates structured CoT reasoning. Based on these COTreasonig, we then use Qwen2.5-14B to generate concise answers from structured CoT-style reasoning. DeepSeek-V3 is subsequently employed to review the reasoning-answer pairs, ensuring logical consistency and semantic alignment.
\begin{figure}[t!]
  \begin{center}
    \includegraphics[width=1\textwidth]{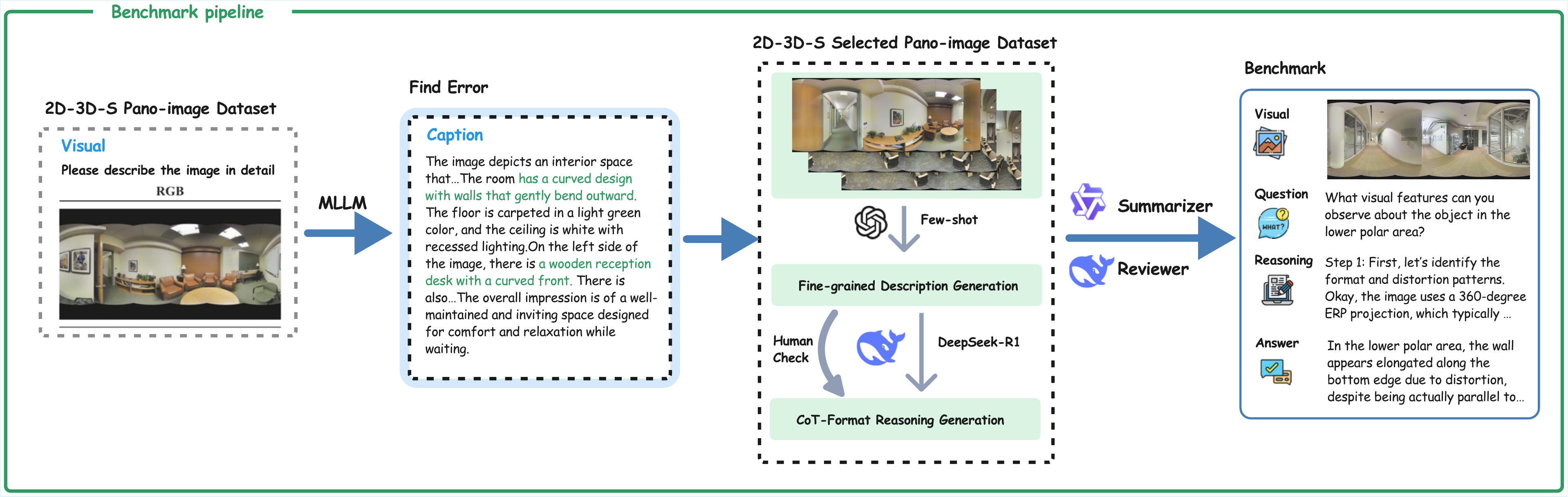}
  \end{center}
  \caption{Benchmark Construction Pipeline.
  }
  \vspace{-12pt}
  \label{fig:Benchmark}
\end{figure}

\begin{table}[t!]
\centering
\caption{Test existing SOTA MLLM on our OmniVQA Bench}
\label{tab:metrics}
\renewcommand{\tabcolsep}{2pt}
\resizebox{\textwidth}{!}{%
\begin{tabular}{l *{12}{c}}
\toprule
\multirow{2}{*}{\textbf{Model}} & \multicolumn{3}{c}{\textbf{BERTScore}~\cite{bertscore}$\uparrow$} & \multicolumn{3}{c}{\textbf{SBERTScore}~\cite{sentencebert}$\uparrow$} & \multicolumn{3}{c}{\textbf{QwenScore}$\uparrow$} & \multicolumn{3}{c}{\textbf{DeepSeekScore}$\uparrow$} \\
\cmidrule(lr){2-4} \cmidrule(lr){5-7} \cmidrule(lr){8-10} \cmidrule(lr){11-13}
 & R & A & F1 & R & A & F1 & R & A & F1 & R & A & F1 \\
\midrule
\multicolumn{13}{l}{\textbf{LlavaOnevision Series~\cite{llava-onevision}}} \\
LlavaOnevision Qwen2-7b-ov & 0.8353 & 0.8837 & 0.8588 & 0.4973 & 0.5802 & 0.5355 & 0.2670 & 0.2470 & 0.2566 & 0.2336 & 0.2509 & 0.2419 \\
LlavaOnevision Qwen2-7b-si & 0.8218 & 0.8809 & 0.8503 & 0.4758 & 0.5661 & 0.5170 & 0.2400 & 0.2390 & 0.2395 & 0.1714 & 0.2533 & 0.2044 \\
\midrule
\multicolumn{13}{l}{\textbf{Janus Series~\cite{wu2024janus, chen2025januspro}}} \\
JanusPro-7B           & 0.8395 & 0.8890 & 0.8635 & \underline{0.5401} & 0.6379 & \underline{0.5849} & 0.3420 & 0.3170 & 0.3290 & 0.3262  & 0.3614 & 0.3429 \\
JanusPro-1B           & 0.6788 & 0.8720 & 0.7633 & 0.4079 & 0.5122 & 0.4541 & 0.2500 & 0.2160 & 0.2318 & 0.1900 & 0.3240 & 0.2395 \\
Janus-1.3B             & 0.8278 & 0.8836 & 0.8548 & 0.4716 & 0.6007 & 0.5284 & 0.2450 & 0.2470 & 0.2460 & 0.2119 & 0.2468 & 0.2281 \\
\midrule
\multicolumn{13}{l}{\textbf{InternVL Series~\cite{Internvl2_5}}} \\
InternVL2\_5-2B        & 0.8418 & \underline{0.8921} & 0.8662 & 0.4822 & 0.6150 & 0.5406 & 0.2840 & 0.2990 & 0.2913 & 0.2961 & 0.3246 & 0.3203 \\
InternVL2\_5-4B        & 0.8384 & 0.8892 & 0.8630 & 0.5124 & 0.6230 & 0.5623 & 0.3140 & 0.3140 & 0.3140 & 0.2289 & 0.2972 & 0.3029 \\
InternVL2\_5-8B        & 0.8388 & 0.8898 & 0.8636 & 0.4932 & 0.6218 & 0.5500 & 0.3250 & 0.3190 & 0.3220 & 0.3124 & 0.3403 & 0.3363 \\
\midrule
\multicolumn{13}{l}{\textbf{DeepSeek Series~\cite{lu2024deepseekvl,wu2024deepseekvl2}}} \\
DeepSeekVL2-small       & 0.8402 & \textbf{0.8940} & \underline{0.8663} & 0.4707 & 0.6252 & 0.5370 & 0.3350 & 0.3390 & 0.3370 & \underline{0.4102} & \underline{0.4554} & \underline{0.4316} \\
DeepSeekVL-3b-chat      & 0.8325 & 0.8749 & 0.8532 & 0.4312 & 0.5100 & 0.4673 & 0.2260 & 0.2130 & 0.2193 & 0.2164 & 0.2889 & 0.2476 \\
DeepSeekVL-1.3b-chat    & 0.8177 & 0.8629 & 0.8397 & 0.2866 & 0.4246 & 0.3422 & 0.1070 & 0.1100 & 0.1085 & 0.1050 & 0.1255 & 0.1143 \\ 
\midrule
\multicolumn{13}{l}{\textbf{Qwen Series~\cite{wang2024qwen2vl, ref23}}} \\ 
QwenVL2-2B-Instruct   & \underline{0.8420} & 0.8919 & \underline{0.8663} & 0.4800 & 0.6373 & 0.5476 & 0.2770 & 0.2910 & 0.2838 & 0.3067 & 0.3456 & 0.3461 \\
QwenVL2-7B-Instruct    & 0.8421 & 0.8917 & 0.8662 & 0.4812 & 0.6341 & 0.5472 & 0.3080 & 0.3050 & 0.3065 & 0.3240 & 0.3711 & 0.3675 \\
QwenVL2.5-3B-Instruct   & 0.8391 & 0.8908 & 0.8642 & 0.5083 & \underline{0.6549} & 0.5724 & \underline{0.3500} & \underline{0.3530} & \underline{0.3515} & 0.3625 & 0.4008 & 0.3966 \\
QwenVL2.5-7B-Instruct   & \textbf{0.8444} & 0.8920 & \textbf{0.8675} & \textbf{0.5870} & \textbf{0.6687} & \textbf{0.6252} & \textbf{0.4400} & \textbf{0.4220} & \textbf{0.4308} & \textbf{0.4638} & \textbf{0.5555} & \textbf{0.5055} \\
\bottomrule
\end{tabular}%
}
\vspace{-16pt}
\end{table}

\section{360-R1 Framework}

To enhance multi-modal reasoning capabilities in omnidirectional environments, we propose a rule-based reinforcement learning post-training strategy. Given that Qwen2.5-VL-Instruct~\cite{ref23} has already undergone extensive instruction tuning on large-scale datasets—ensuring stable, compliant generation—we directly apply rule-based RL on this model to promote structured reasoning and enforce output validity.

\subsection{Reward Function Design}

We design three reward functions to guide reinforcement learning using Group Relative Policy Optimization. These rewards aim to improve the quality of reasoning, the accuracy of answers, and the consistency of output formatting, thus enhancing the model's generalization and robustness.

\noindent \textbf{Reasoning Process Similarity Reward} evaluates the semantic and logical alignment between the generated reasoning and a reference COT reasoning. It extracts reasoning segments enclosed within predefined tags and computes a similarity score using the DeepSeek-V3~\cite{deepseekai2025deepseekv3technicalreport} model via prompt-based evaluation. The score, ranging from 0.0 to 1.0, is tolerant to surface-level variations while sensitive to logical discrepancies, providing fine-grained feedback to guide reasoning generation.

\noindent \textbf{Answer Semantic Accuracy Reward} measures the semantic similarity between the generated and reference answers. Answer content is automatically extracted from the structured output, and a similarity score is computed using DeepSeek-V3~\cite{deepseekai2025deepseekv3technicalreport}. The reward encourages the generation of semantically correct responses, even when lexical or syntactic differences exist.

\noindent \textbf{Structured Format Compliance Reward.}  
This binary reward (1.0 or 0.0) checks whether the output adheres to a predefined structured format. It verifies the presence, order, and correct nesting of essential components such as reasoning and answer sections. 

\subsection{Group Relative Policy Optimization}
We adopt Group Relative Policy Optimization (GRPO) to achieve stable policy updates while effectively incorporating rule-based rewards. Unlike standard Proximal Policy Optimization (PPO), GRPO eliminates the need for a separate value function by estimating the advantage through group-wise normalized rewards.
For each input question \( q \), GRPO samples a group of \( G \) responses \( \{o_1, \ldots, o_G\} \) from the old policy \( \pi_{\text{old}} \). A reward model assigns scalar scores \( \{r_1, \ldots, r_G\} \), which are then normalized within the group:
\begin{equation}
\setlength{\abovedisplayskip}{3pt}
\setlength{\belowdisplayskip}{3pt}
\hat{r}_i = \frac{r_i - \text{mean}(r)}{\text{std}(r)},
\end{equation}
and used as the advantage for all tokens in the \( i \)-th response:
\begin{equation}
\setlength{\abovedisplayskip}{3pt}
\setlength{\belowdisplayskip}{3pt}
\hat{A}_{i,t} = \hat{r}_i.
\end{equation}
The clipped surrogate loss is computed as:
\begin{equation}
\setlength{\abovedisplayskip}{3pt}
\setlength{\belowdisplayskip}{3pt}
\mathcal{L}_{\text{clip}} = -\mathbb{E} \left[ \min \left( r_t \cdot \hat{A}_t, \text{clip}(r_t, 1 - \epsilon, 1 + \epsilon) \cdot \hat{A}_t \right) \right],
\end{equation}
where \( r_t = \frac{\pi_\theta(o_t|q, o_{<t})}{\pi_{\text{old}}(o_t|q, o_{<t})} \) is the token-level probability ratio.
To prevent the updated policy from deviating too far from the reference policy, a KL divergence regularization term is included:
\begin{equation}
\setlength{\abovedisplayskip}{3pt}
\setlength{\belowdisplayskip}{3pt}
\mathcal{L}_{\text{GRPO}}(\theta) = \mathcal{L}_{\text{clip}} + \beta \cdot \text{KL}(\pi_\theta \| \pi_{\text{ref}}).
\end{equation}
This group-relative formulation aligns well with reward models trained on preference data, and avoids the computational burden of training a separate value network. The combination of group-based advantage, clipping, and KL regularization ensures stable and efficient policy updates under rule-based supervision.

\section{Experiment}
\noindent \textbf{Experimental Setup.} We conduct our experiments on OmniVQABench for evaluation and use the OmniVQA dataset for training. The base model is Qwen2.5-VL-7B-Instruct. We compare our method against several state-of-the-art vision-language models, including the Qwen Series (QwenVL2-2B to QwenVL2.5-7B)\cite{wang2024qwen2vl, ref23}, LlavaOnevision Series (Qwen2-7B-OV, Qwen2-7B-SI)\cite{llava-onevision}, Janus Series (1B to 7B)\cite{wu2024janus, chen2025januspro}, InternVL2.5 Series (2B to 8B)\cite{Internvl2_5}, and DeepSeekVL Series (small to 7B)~\cite{lu2024deepseekvl,wu2024deepseekvl2}. Our training strategy employs reinforcement learning with Group Relative Policy Optimization (GRPO).All RL experiments are performed on 4×H800 GPUs for 550 steps (over 1 epochs), taking approximately 2.8 days. Full hyperparameter settings and training scripts are provided in the supplementary material.

Performance is evaluated using four metrics: BERTScore~\cite{bertscore}, SentenceBERTScore~\cite{sentencebert}, QwenScore, and DeepSeekScore. BERTScore and SentenceBERTScore measure semantic similarity at token and sentence level, respectively. QwenScore and DeepSeekScore assess both reasoning quality and answer accuracy via semantic alignment with human annotations. QwenScore uses the QwenLLM~\cite{qwen2025qwen25technicalreport} with structured prompting to rate both reasoning and answer quality, combining them into an F1-based score. DeepSeekScore uses DeepSeek-chat~\cite{deepseekai2025deepseekv3technicalreport} to output a score between 0.0 and 1.0. For LLM-based metrics, the final evaluation includes individual reasoning and answer scores as well as their harmonic mean, to comprehensively reflect multimodal reasoning and semantic consistency.

\begin{wraptable}{r}{9.5cm}
\centering
\caption{Comparison between 360-R1 and QwenVL2.5-7B-Instruct.}
\label{tab:model_comparison}
\renewcommand{\tabcolsep}{3.3pt}
\resizebox{\linewidth}{!}{
\begin{tabular}{l *{6}{c}}
\toprule
\multirow{2}{*}{\textbf{Model}} & \multicolumn{3}{c}{\textbf{BERTScore~\cite{bertscore}}$\uparrow$} & \multicolumn{3}{c}{\textbf{SentenceBERTScore~\cite{sentencebert}}$\uparrow$} \\
\cmidrule(lr){2-4} \cmidrule(lr){5-7}
 & R & A & F1 & R & A & F1 \\
\midrule
QwenVL2.5-7B-Instruct & 0.8444 & 0.8920 & 0.8675 & 0.5870 & 0.6687 & 0.6252 \\
360-R1 & \textbf{0.8722} & \textbf{0.8959} & \textbf{0.8839} & \textbf{0.6100} & \textbf{0.6705} & \textbf{0.6388} \\
\midrule
\textbf{Improvement} & \textcolor{red}{\textbf{+2.78\%}} & \textcolor{red}{\textbf{+0.39\%}} & \textcolor{red}{\textbf{+1.64\%}} & \textcolor{red}{\textbf{+2.30\%}} & \textcolor{red}{\textbf{+0.18\%}} & \textcolor{red}{\textbf{+1.36\%}} \\
\bottomrule
\end{tabular}}
\vspace{0.5cm}
\renewcommand{\tabcolsep}{3.5pt}
\resizebox{\linewidth}{!}{
\begin{tabular}{l *{6}{c}}
\toprule
\multirow{2}{*}{\textbf{Model}} & \multicolumn{3}{c}{\textbf{QwenScore}$\uparrow$} & \multicolumn{3}{c}{\textbf{DeepSeekScore}$\uparrow$} \\
\cmidrule(lr){2-4} \cmidrule(lr){5-7}
 & R & A & F1 & R & A & F1 \\
\midrule
QwenVL2.5-7B-Instruct & 0.4400 & 0.4220 & 0.4308 & 0.4638 & 0.5555 & 0.5055 \\
360-R1 & \textbf{0.4860} & \textbf{0.5140} & \textbf{0.4996} & \textbf{0.4847} & \textbf{0.6282} & \textbf{0.5472} \\
\midrule
\textbf{Improvement} & \textcolor{red}{\textbf{+4.60\%}} & \textcolor{red}{\textbf{+9.20\%}} & \textcolor{red}{\textbf{+6.88\%}} & \textcolor{red}{\textbf{+2.09\%}} & \textcolor{red}{\textbf{+7.27\%}} & \textcolor{red}{\textbf{+4.17\%}} \\
\bottomrule
\end{tabular}}
\vspace{-20pt}
\end{wraptable}

\subsection{Results Anlysis}

Compared to the baseline QwenVL2.5-7B-Instruct, our 360-R1 model demonstrates consistent improvements across all evaluation metrics. As shown in Table~\ref{tab:model_comparison}, 360-R1 achieves the highest F1 scores, with notable gains in QwenScore (+6.88\%) and DeepSeekScore (+4.17\%). Specifically, the reasoning sub-score of QwenScore increases by 4.60\%, while the answer quality improves by 9.20\%, reflecting enhanced logical interpretation and factual correctness. SentenceBERTScore and BERTScore also exhibit steady improvements (+1.36\% and +1.64\%, respectively), indicating better semantic fluency and lexical alignment.

We attribute these improvement to two key factors: the use of GRPO and the OmniVQA dataset. GRPO enables the model to learn from relative quality comparisons across answer groups, effectively optimizing for both reasoning soundness and answer precision. It provides a structured reward signal guided by multi-domain examples and rating scales, enhancing the model’s ability to align with human judgment. Meanwhile, the OmniVQA dataset offers a rich set of vision-language queries covering diverse and complex reasoning types, such as spatial understanding and object relations. Together with our scoring-guided training pipeline, these elements equip the model with robust multimodal reasoning and semantic alignment capabilities. Overall, the results validate the effectiveness of our training approach in improving both explicit reasoning quality and implicit semantic fidelity.

\subsection{Ablation Study}
To better understand the effects of model initialization and reward design in our GRPO-based training framework, we conduct ablation studies focusing on two aspects: (1) comparing base models with different parameter scales, and (2) analyzing the impact of reward weight configurations. All models are fine-tuned using GRPO for 200 steps to ensure fair comparison under limited training conditions.

\noindent \textbf{Impact of Model Parameter Scale}  
As shown in Table~\ref{tab:model_param_comparison}, we compare two variants: 360-R1-3B and 360-R1-7B, trained based on QwenVL2.5-3B-Instruct and QwenVL2.5-7B-Instruct respectively. Despite undergoing the same training steps, the larger 7B model consistently outperforms its 3B counterpart across all metrics. QwenScore-F1 improves by 7.94\%, and DeepSeekScore-F1 by 8.07\%, demonstrating the benefits of scaling in pre-trained vision-language models. These gains are particularly notable in reasoning-intensive metrics, indicating enhanced generalization capability and stronger prior knowledge in the larger model.

\noindent \textbf{Impact of Reward Weight Configuration}  
We further investigate how different reward weightings affect model performance. Table~\ref{tab:reward_weight_ablation} compares three reward configurations in the format Base:Reasoning:Answer, where the base reward is fixed at 0.1 and the remaining weights control the emphasis between reasoning and answer accuracy. The default setting is 0.1:0.45:0.45. Emphasizing answer accuracy (0.1:0.4:0.5) leads to a slight improvement in SentenceBERTScore-F1 (+0.82\%), but results in a decline in QwenScore-F1 (–7.00\%), suggesting weakened logical quality. In contrast, prioritizing reasoning (0.1:0.5:0.4) increases DeepSeekScore-F1 (+1.29\%), but lowers SentenceBERTScore-F1. These results highlight a trade-off between reasoning and answer fidelity, and demonstrate the GRPO framework’s sensitivity to reward design. The balanced configuration (0.1:0.45:0.45) offers the most stable and competitive performance and is therefore adopted as the default in our final model.

\begin{table}[htbp]
  \centering
  \begin{minipage}[t]{0.46\textwidth}
    \centering
    \caption{Comparison of Model Parameters (trained for 200 steps).}
    \label{tab:model_param_comparison}
    \renewcommand{\tabcolsep}{1pt}
    \resizebox{\linewidth}{!}{
      \begin{tabular}{lcc}
        \toprule
        \textbf{Evaluation Metric} & \textbf{360-R1-3B} & \textbf{360-R1-7B} \\
        \midrule
        BERTScore-F1 & 0.8556 & 0.8610 \\
        SentenceBERTScore-F1 & 0.5872 & 0.6324 \\
        QwenScore-F1 & 0.3708 & 0.4502 \\
        DeepSeekScore-F1 & 0.4292 & 0.5099 \\
        \bottomrule
      \end{tabular}
    }
  \end{minipage}%
  \hfill
  \begin{minipage}[t]{0.50\textwidth}
    \centering
    \caption{Comparison of Different Reward Weight Configurations (360-R1-7B, trained for 200 steps).}
    \label{tab:reward_weight_ablation}
    \renewcommand{\tabcolsep}{2pt}
    \resizebox{\linewidth}{!}{
      \begin{tabular}{lccc}
        \toprule
        \multirow{2}{*}{\textbf{Evaluation Metric}} & \multicolumn{3}{c}{\textbf{Format:Reasoning:Answer}} \\
        \cmidrule(lr){2-4}
        & \textbf{0.1:0.45:0.45} & \textbf{0.1:0.4:0.5} & \textbf{0.1:0.5:0.4} \\
        \midrule
        BERTScore-F1 & 0.8610 & 0.8606 & 0.8616 \\
        SentenceBERTScore-F1 & 0.6324 & 0.6376 & 0.6261 \\
        QwenScore-F1 & 0.4502 & 0.4187 & 0.4295 \\
        DeepSeekScore-F1 & 0.5099 & 0.4992 & 0.5165 \\
        \bottomrule
      \end{tabular}
    }
  \end{minipage}
\end{table}

\section{Conclusion, Limitations, and Future Work}
We introduce OmniVQA, the first omnidirectional VQA dataset, and the corresponding OmniVQABench, addressing key challenges in object identification, attribute analysis, and spatial reasoning within 360° panoramic images. Our proposed 360-R1 method leverages rule-based reinforcement learning with GRPO optimization, significantly improving reasoning accuracy and semantic consistency in omnidirectional contexts. ablation studies further validate the effectiveness of both the proposed structured reward functions and the GRPO optimization approach.

\noindent \textbf{Limitations}
Despite notable advancements, our work has certain limitations. The OmniVQA dataset is limited in scale (approximately 4.8K QA pairs) and currently covers only indoor panoramic scenes, restricting generalization to broader contexts. Additionally, our GRPO-based reinforcement learning framework requires substantial computational resources, potentially limiting accessibility and reproducibility for researchers with constrained computational capabilities.

\noindent \textbf{Future Work}
Future research directions include expanding OmniVQA to diverse scenarios such as outdoor environments and dynamic scenes, enhancing model generalization. Additionally, we aim to develop more adaptive and computationally efficient reward mechanisms and optimization strategies. Finally, integrating additional modalities—such as audio cues, temporal information, and embodied interactions—can further enrich omnidirectional reasoning capabilities, supporting practical real-world applications.

\noindent \textbf{Broader Societal Impact} 
OmniVQA and the 360-R1 framework provide important progress for vision-language understanding in omnidirectional environments, enabling new applications in robotics, assistive technology, immersive education, and virtual reality. However, the deployment of such systems may raise concerns regarding user privacy, surveillance, and potential misuse in security-critical scenarios. Dataset limitations in terms of cultural and environmental diversity may also introduce unintended biases, affecting model fairness and generalizability. We encourage future research to emphasize responsible data collection, fairness evaluation, and transparent reporting to mitigate these risks and maximize societal benefit.

\clearpage
\bibliographystyle{plain}  
\bibliography{ref1}

\tcbset{
  mybox/.style={
    colback=white,
    colframe=black,
    boxrule=0.6pt,
    arc=1mm,
    leftrule=0.6pt,
    rightrule=0.6pt,
    toptitle=1mm,
    bottomtitle=1mm,
    fonttitle=\bfseries,
    coltitle=black
  },
  torchbox/.style={
    colback=white,
    colframe=gray!70,
    boxrule=0.6pt,
    arc=1mm,
    leftrule=0.6pt,
    rightrule=0.6pt,
    toptitle=1mm,
    bottomtitle=1mm,
    fonttitle=\bfseries,
    coltitle=black
  }
}

\lstset{
  basicstyle=\ttfamily\small,
  breaklines=true,
  columns=fullflexible
}

\clearpage
\appendix
\renewcommand\thesection{\Alph{section}}
\renewcommand\thesubsection{\thesection.\arabic{subsection}}

\section*{Appendix}
\addcontentsline{toc}{section}{Roadmap of Appendix}
\section{Prompt Templates}
\label{app:prompt-templates}
This part collects all of the system and user prompts we use to drive our VQA pipeline. Each prompt is encapsulated in a styled box for clarity, and includes guidance on inputs, expected outputs, and formatting conventions.

\subsection{Fine-Grained Description Generation (Perception Stage)}
\noindent\textbf{Role in Experiment:}  
This prompt is used to elicit a step-by-step, fine-grained analytical description of a panoramic image from a LoRA-fine-tuned \texttt{Qwen2.5VL-7B-Instruct} model. It provides the initial perception stage input for subsequent reasoning and answer generation modules. Detailed spatial relationship language is required to mitigate distortion-related errors and support downstream Chain-of-Thought (CoT) reasoning.

\begin{tcolorbox}[mybox,title={A.1 Fine-Grained Description Generation}]
\textbf{Input:} Image and natural-language question

\begin{lstlisting}
Please generate a fine-grained analytical description based on the image
and question provided below. Clearly present the complete thought process
rather than merely providing a final answer. When the question involves
spatial relationships, explicitly describe them using clear directional
terms such as 'up', 'down', 'left', 'right', 'front', 'back', 'covering',
or 'adjacent to'.
\end{lstlisting}
\end{tcolorbox}

\subsection{Chain-of-Thought (CoT) Reformatting}
\noindent\textbf{Role in Experiment:}  
This prompt is applied to convert free-form descriptive reasoning into a structured, multi-step Chain of Thought using the \texttt{DeepSeek-R1} model. It enforces explicit logical progression, which facilitates downstream evaluation, answer summarization, and reasoning-based reward functions.

\begin{tcolorbox}[mybox,title={A.2 Chain-of-Thought Reformatting}]
\textbf{System:}
\begin{lstlisting}
You are a Chain of Thought (CoT) reformatting expert. Your task is to
transform descriptive reasoning into natural, step-by-step thinking
that reflects how one would logically work through a problem. Ensure
each step builds on the previous one, retains all critical information,
and concludes with a direct answer to the original question.
\end{lstlisting}

\textbf{User:}
\begin{lstlisting}
Guidelines: 
1. Structure content into 3-5 explicit reasoning steps.
2. Use transition words (e.g., "First," "Next," "Then") and thinking-style
   phrases (e.g., "Hmm," "Let's see," "Okay").
3. Ensure logical flow: each step derives from the prior one.
4. Preserve all original information.
5. End with a clear conclusion.

Question: {question}
Original reasoning: {reasoning}

Transformed COT reasoning:
\end{lstlisting}
\end{tcolorbox}

\subsection{CoT Summarisation (Answer Synthesis)}
\noindent\textbf{Role in Experiment:}  
This prompt guides the \texttt{Qwen2.5-14B-Instruct} model to convert a structured Chain of Thought into a concise, factual answer. It ensures that the final answer is directly grounded in the provided reasoning context, preventing hallucination or unsupported information.

\begin{tcolorbox}[mybox,title={A.3 CoT-to-Answer Summarisation}]
\begin{lstlisting}
{"role":"system","content":"You are a precision-focused assistant. Generate concise
and factual answers based on the given question and reasoning context.
Answer should directly address the question using ONLY information from the reasoning."},
{"role":"user","content":prompt}
\end{lstlisting}
\end{tcolorbox}

\subsection{Answer Consistency Review (VQA Evaluation)}
\noindent\textbf{Role in Experiment:}  
This prompt uses \texttt{DeepSeek-V3} to evaluate whether a given answer is consistent with the original question and structured reasoning. It is critical for post-hoc filtering, reward calculation during RL post-training, and ensuring dataset quality.

\begin{tcolorbox}[mybox,title={A.4 Answer Consistency Review}]
\begin{lstlisting}
{"role":"system","content":"You are an expert reviewer in the Visual Question
Answering (VQA) domain. Your task is to evaluate whether the provided answer
is consistent with the question and the structured reasoning.

Evaluation criteria:
1. Factual Accuracy
2. Logical Consistency
3. Completeness

If errors exist, provide a corrected version. If fully consistent, reply only:
CONSISTENT."},
{"role":"user","content":"Question: {question}
Reasoning: {reasoning}
Current Answer: {answer}

Please review item #{index}.
If not consistent, provide corrected answer. If consistent, reply: CONSISTENT."}
\end{lstlisting}
\end{tcolorbox}

\subsection{Quality Scoring (Reference Comparison)}
\noindent\textbf{Role in Experiment:}  
This prompt instructs the Qwen LLM to assign discrete scores to the generated reasoning and answer by comparing them with high-quality references. The output scores are used for supervised training, evaluation, and as components in reward functions for RL-based post-training.

\begin{tcolorbox}[mybox,title={A.5 Reference-Based Quality Scoring}]
\begin{lstlisting}
You are a text similarity evaluator. Your task is to compare a candidate's
reasoning and answer with reference versions. Rate each on a scale of
1-5 (1 = completely incorrect, 5 = completely correct).
IMPORTANT: Respond ONLY with two lines:
REASONING_SCORE: [1/2/3/4/5]
ANSWER_SCORE:   [1/2/3/4/5]

---  

Question: {question}
Reference reasoning: {reference_reasoning}
Candidate reasoning: {candidate_reasoning}
Reference answer:   {reference_answer}
Candidate answer:   {candidate_answer}
\end{lstlisting}
\end{tcolorbox}

\subsection{Semantic Consistency Scoring}
\noindent\textbf{Role in Experiment:}  
This prompt queries \texttt{DeepSeek-V3} to compute a fine-grained semantic similarity score (range: 0.0000–1.0000) between reference and candidate answers. This score is used in both dataset filtering (e.g., during iterative refinement) and as a reward signal in RL training.

\begin{tcolorbox}[mybox,title={A.6 Semantic Consistency Scoring}]
\begin{lstlisting}
{"role":"system","content":"You are a professional evaluation assistant.
Analyze the semantic consistency between reference and candidate answers.
Guidelines:
1. Score range: 0.0-1.0
2. Consider accuracy and context
3. Return ONLY the numeric score with 4 decimal places
4. No additional text."},
{"role":"user","content":"Reference: {reference}\nCandidate: {candidate}\nScore:"}
\end{lstlisting}
\end{tcolorbox}

\subsection{Reasoning Similarity Scoring (GRPO Reward)}
\noindent\textbf{Role in Experiment:}  
This prompt is used during Group Relative Policy Optimization (GRPO) training to compute the semantic similarity between the generated and reference reasoning chains, using \texttt{DeepSeek-Chat}. The returned score is directly used as the Reasoning Process Similarity Reward.

\begin{tcolorbox}[mybox,title={A.7 Reasoning Similarity Scoring Prompt}]
\textbf{System:}
\begin{lstlisting}
You are an AI assistant that evaluates the similarity between two reasoning processes. Your response MUST follow the format 'Similarity score: <score>', where <score> is a single floating-point number between 0.0 and 1.0 representing the similarity score.
\end{lstlisting}

\textbf{User:}
\begin{lstlisting}
Compare these two reasoning passages and return a similarity score 0-1.

Generated:
{gen_text}

Reference:
{ref_text}
\end{lstlisting}
\end{tcolorbox}

\subsection{Answer Similarity Scoring (GRPO Reward)}
\noindent\textbf{Role in Experiment:}  
This prompt is used during GRPO training to assess the semantic similarity between the generated answer and the reference answer, using \texttt{DeepSeek-Chat}. The score forms the Answer Semantic Accuracy Reward.

\begin{tcolorbox}[mybox,title={A.8 Answer Similarity Scoring Prompt}]
\textbf{System:}
\begin{lstlisting}
You are an AI assistant that evaluates the similarity between two answer. Your response MUST follow the format 'Similarity score: <score>', where <score> is a single floating-point number between 0.0 and 1.0 representing the similarity score.
\end{lstlisting}

\textbf{User:}
\begin{lstlisting}
Compare these two answer texts and return a similarity score 0-1.

Generated:
{gen_text}

Reference:
{ref_text}
\end{lstlisting}
\end{tcolorbox}

\subsection{Fine-Grained Description Prompt for GPT-4o (Few-Shot)}

\noindent
\textbf{Role in Experiment:}  
This prompt is used to guide GPT-4o in generating high-quality, fine-grained descriptions of 360-degree panoramic images. The prompt provides multiple few-shot examples, each consisting of a panoramic image and a detailed description that explicitly addresses distortion effects and spatial relationships.

\begin{tcolorbox}[mybox,title={A.9 Few-Shot Prompt for Fine-Grained Description (GPT-4o)}]
\textbf{System:}
\begin{lstlisting}
You are an expert visual-language model that generates fine-grained descriptions of 360-degree panoramic indoor scenes. Your goal is to identify furniture, materials, lighting, and spatial layout in natural and coherent English.  
\end{lstlisting}

\textbf{User (Few-shot):} 
\begin{lstlisting}
Please generate a paragraph that accurately describes the scene shown in the image. Follow the style of the examples provided, focusing on object identities, spatial relationships, and the contextual organization of the scene. Be as specific and spatially precise as possible.
\end{lstlisting}

\end{tcolorbox}
\vspace{-10em}

\begin{figure}[H]
    \centering
    \includegraphics[width=0.75\linewidth]{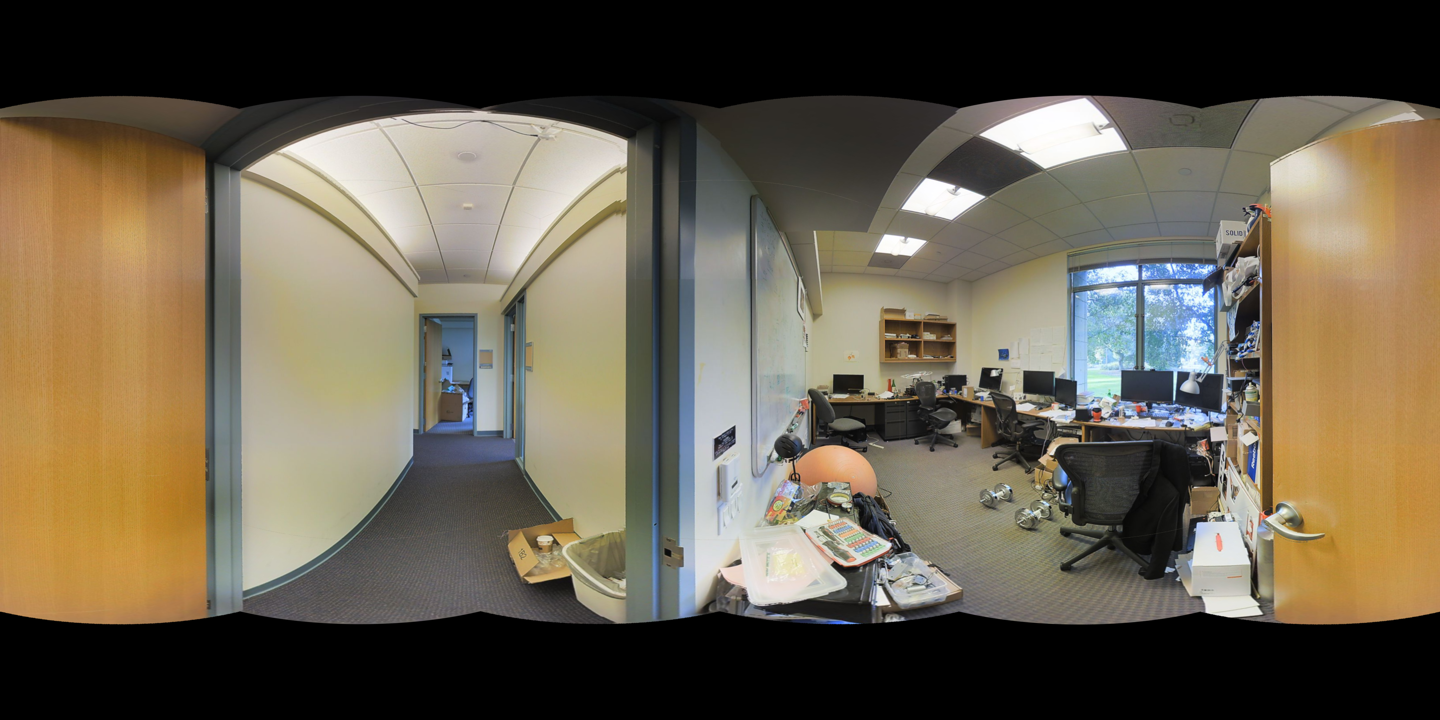}
    \caption*{\textbf{Example 1 Description:}\\
    Consider the impact of the equirectangular projection (ERP) on the lower polar area. The floor's carpet pattern curves outward due to ERP distortion, appearing unnaturally stretched. Despite this warping, the carpet’s neutral color and repeating design remain distinguishable, with their consistency helping to visually anchor the space. When analyzing furniture placement, the semicircular arrangement of tables and chairs is an illusion—distortion causes three rectangular tables (placed in three directions) to appear curved. Zooming in reveals black-backed chairs around the tables, with papers, staplers, notebooks, and other office supplies scattered on the tables; the actual layout is angular, not rounded. Moving upward from the floor, bookshelves line the walls, filled with books and small decorative items like plants or figurines. Framed certificates and photographs hang nearby, suggesting a workspace for academics or professionals. These details contrast with the distortion but remain sharp enough to convey the space’s purpose.}
\end{figure}

\begin{figure}[H]
    \centering
    \includegraphics[width=0.75\linewidth]{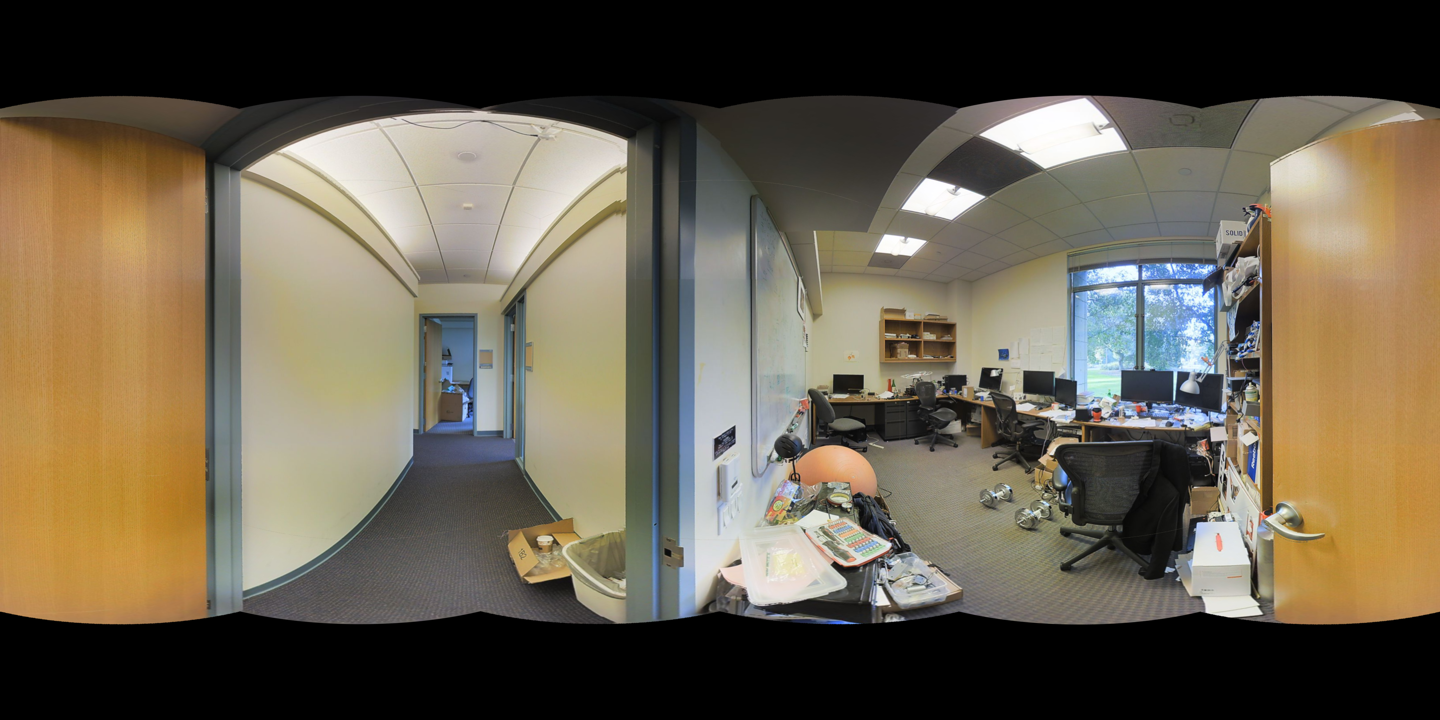}
    \caption*{\textbf{Example 2 Description:}\\
    The image uses an equirectangular projection (ERP), which stretches and distorts objects near the polar regions. In ERP, the top edge represents the "north pole" of the 360° scene, meaning objects near the top may appear warped compared to their real-world arrangement. In the top left polar area, artificial lights are positioned at the junction of the white ceiling and two walls. Due to ERP’s distortion of vertical lines, these lights likely form part of a corner where two walls meet the ceiling in reality. A small wooden board hanging on the side suggests this area is a distinct workspace separated by a wall from the right side. The top right polar area features square recessed ceiling lights and a blackboard with writing; the wall separation implies a different functional zone. Despite distortion, the square recessed lights suggest they are aligned in a straight line on the actual ceiling, while the blackboard’s vertical placement on the wall is evident. The left and right areas are divided by a wall but exist in the same polar region—distortion may make them appear adjacent in the projection, but functionally, they are distinct workspaces. The ceiling lights and boards on both sides serve their respective zones while maintaining proximity for collaboration.}
\end{figure}

\begin{figure}[H]
    \centering
    \includegraphics[width=0.75\linewidth]{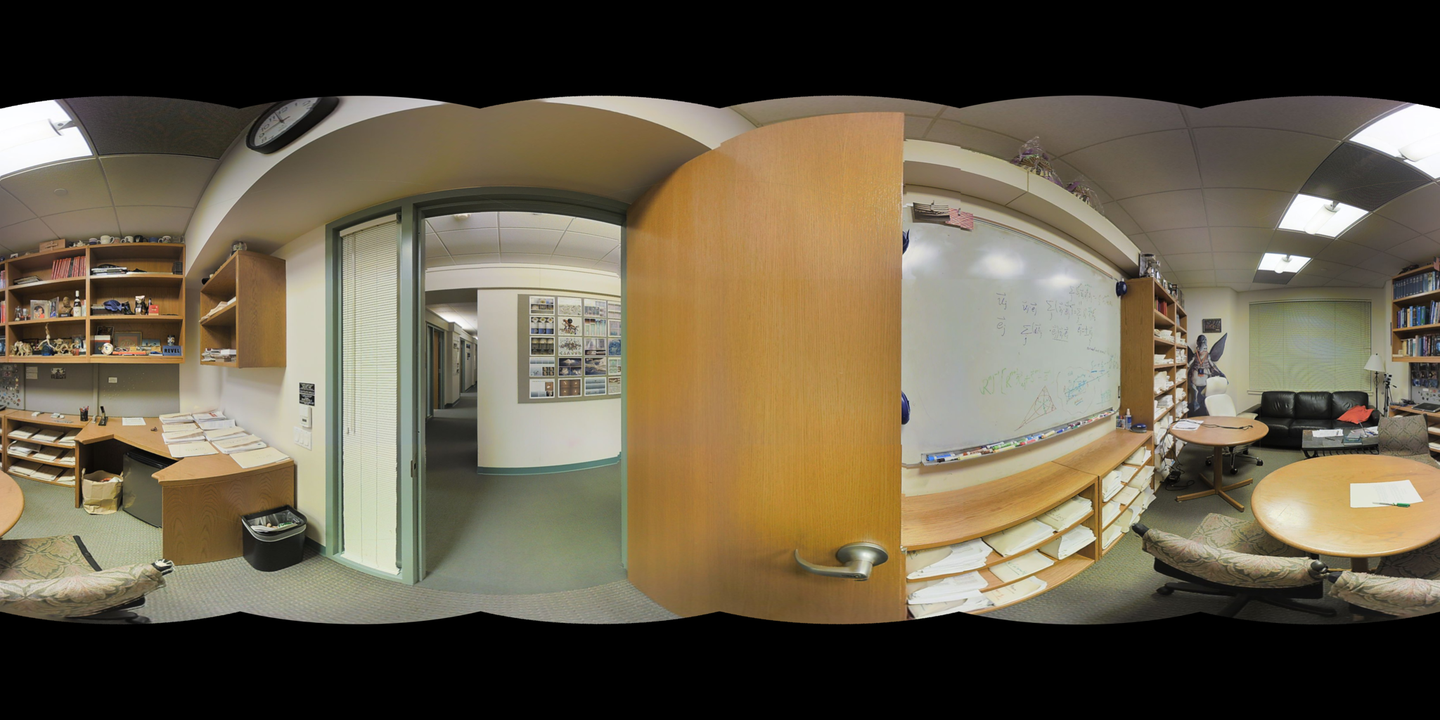}
    \caption*{\textbf{Example 3 Description:}\\
    The scene uses a 360-degree equirectangular projection (ERP), which distorts polar-region objects, causing straight objects like doors to appear curved and the ceiling to look warped. A circular clock near the ceiling remains unobstructed, but ERP stretching complicates perceptions of occlusion. A large door on the right (visually curved due to ERP) partially obscures a whiteboard; their positional relationship persists—if the door is closer to the viewer, it logically blocks the whiteboard behind it. A hanging shelf near the whiteboard also partially blocks items on it. ERP affects visual representation but not physical reality: the door and shelf are positioned to occlude other objects, and their distortion does not erase their spatial relationships. The partial visibility of the whiteboard and shelf items confirms that occlusion occurs despite ERP warping.}
\end{figure}

\begin{figure}[H]
    \centering
    \includegraphics[width=0.75\linewidth]{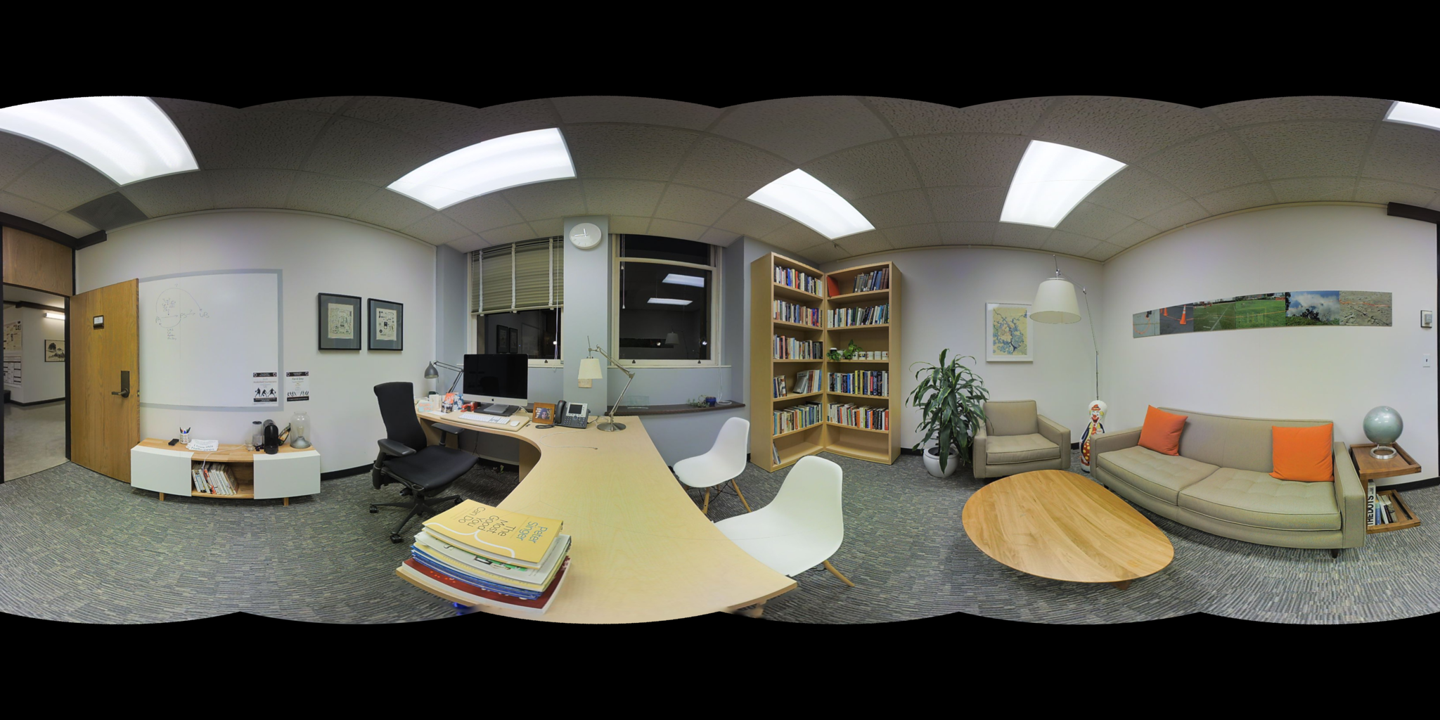}
    \caption*{\textbf{Example 4 Description:}\\
    This is a 360° equirectangular projection (ERP), which distorts polar regions, but does this mean objects do not interact? In the top polar region, artificial lights are "embedded" in the ceiling grid, implying connection or overlap with the grid structure—a clear interaction. In the bottom polar region, books are stacked on a rectangular tabletop (direct physical contact), and an oval coffee table "touches" the carpeted floor. Despite projection-induced distortion of the coffee table’s legs, the contact point (table-to-floor) remains, confirming interaction persists.}
\end{figure}

\begin{figure}[H]
    \centering
    \includegraphics[width=0.75\linewidth]{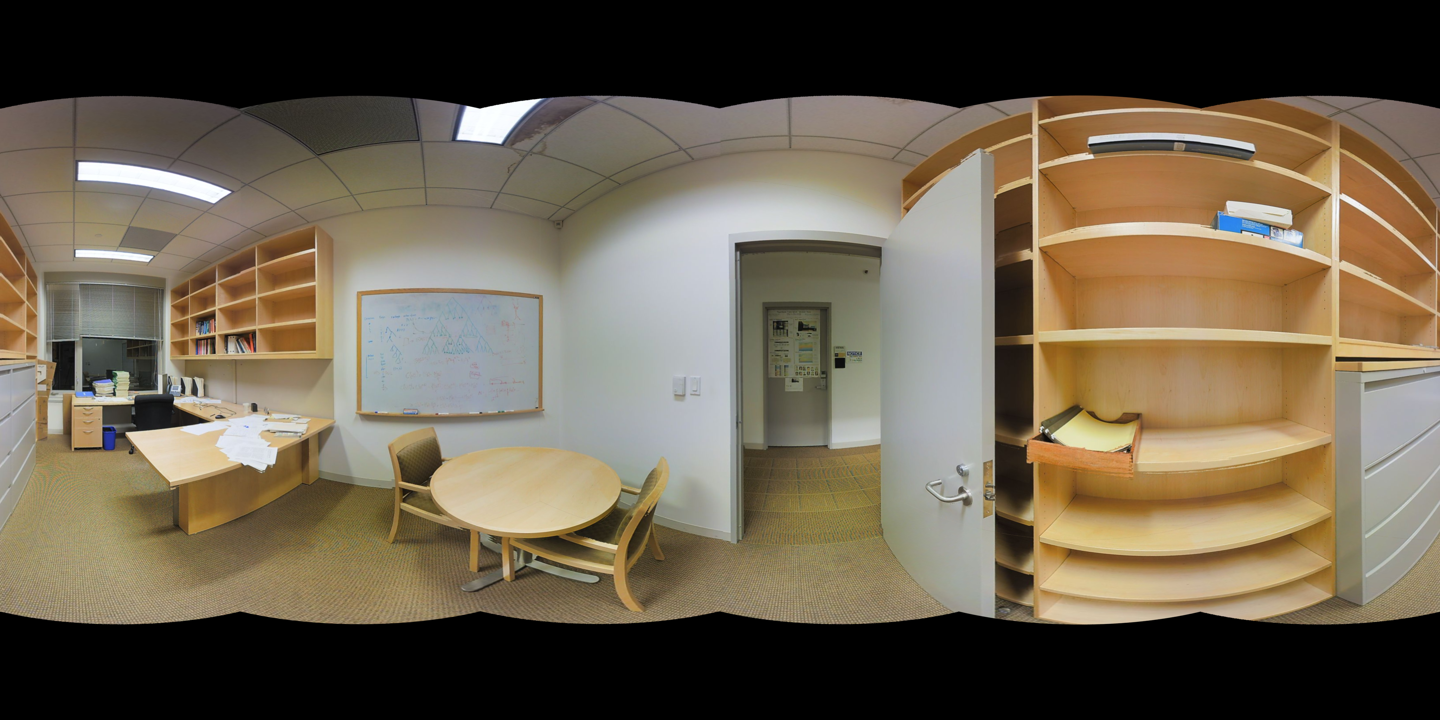}
    \caption*{\textbf{Example 5 Description:}\\
    The image uses an equirectangular projection, which distorts polar regions, making straight objects near the top appear curved or stretched in the projection (though they are physically straight). The upper polar region shows parts of the ceiling, where curvature from distortion affects fluorescent lights (appearing bent), flat ceiling panels in a grid, air vents, and minor ceiling discolorations following the warped pattern. On the right side, a bookshelf extends into this region; its shelves appear slightly warped in the projection but are actually level. The top shelf holds items like a soundbar, books, and storage boxes, whose edges look curved here but are rectangular in real space.}
\end{figure}


\section*{Appendix B: Model Visualization}
\addcontentsline{toc}{section}{Appendix B: Model Visualization}
\vspace{1ex}

\subsection*{B.1 Radar Chart – Reasoning (R)}
\begin{figure}[H]
    \centering
    \includegraphics[width=0.9\linewidth]{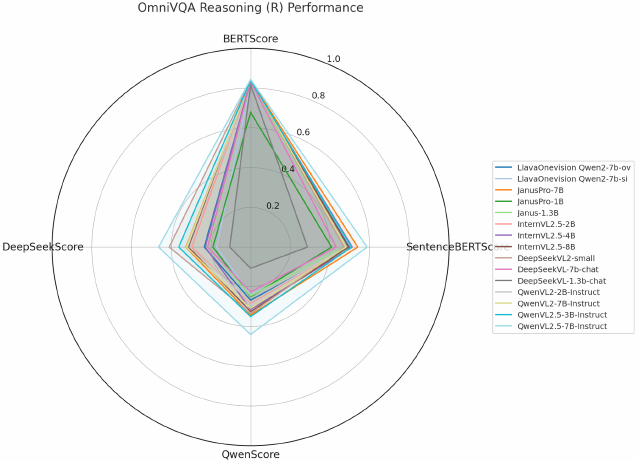}
    \caption{Radar Chart of Reasoning Scores for OmniVQA Benchmark Models.}
    \label{fig:radar-reasoning}
\end{figure}

\noindent
\textbf{Description:}  
This radar chart illustrates each model’s Reasoning scores across the four evaluation metrics (BERTScore, SentenceBERTScore, QwenScore, DeepSeekScore). Each colored polygon corresponds to one model, with values closer to the perimeter indicating higher reasoning performance on that metric. We can see that models like QwenVL2.5-7B-Instruct (sky-blue) occupy a larger area toward the outer edge on all axes, reflecting its strong reasoning quality. In contrast, smaller models (e.g., JanusPro-1B, gray) have polygons drawn closer to the center, especially on metrics like BERTScore and QwenScore, indicating comparatively weaker reasoning performance.

\vspace{2ex}

\subsection*{B.2 Radar Chart – Answer (A)}
\begin{figure}[H]
    \centering
    \includegraphics[width=0.9\linewidth]{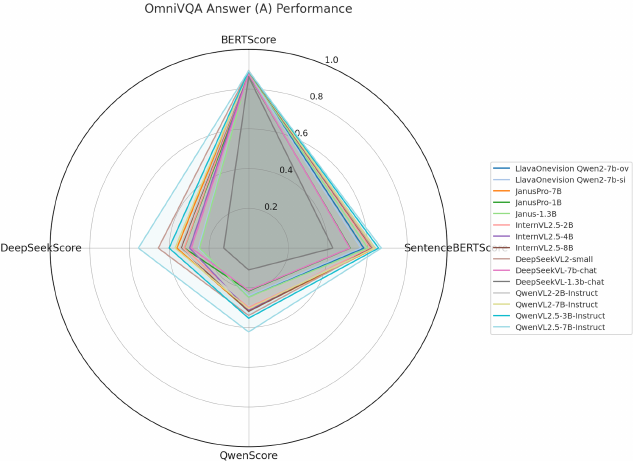}
    \caption{Radar Chart of Answer Accuracy for OmniVQA Benchmark Models.}
    \label{fig:radar-answer}
\end{figure}

\noindent
\textbf{Description:}  
This radar chart shows the Answer accuracy scores for all models on the same four metrics. A broader polygon shape signifies better answer performance across those metrics. Again, QwenVL2.5-7B-Instruct and other larger models form the broadest shapes, indicating superior answer accuracy and semantic alignment. Notably, the DeepSeekVL-1.3b-chat model (black) has a much smaller span, especially on QwenScore and DeepSeekScore axes, highlighting its difficulty in producing high-quality answers compared to the top-performing models. Overall, models that performed well in reasoning tend to also excel in answer accuracy, maintaining consistently higher scores across metrics.

\subsection*{B.3 Radar Chart – Overall F1}
\begin{figure}[H]
    \centering
    \includegraphics[width=0.9\linewidth]{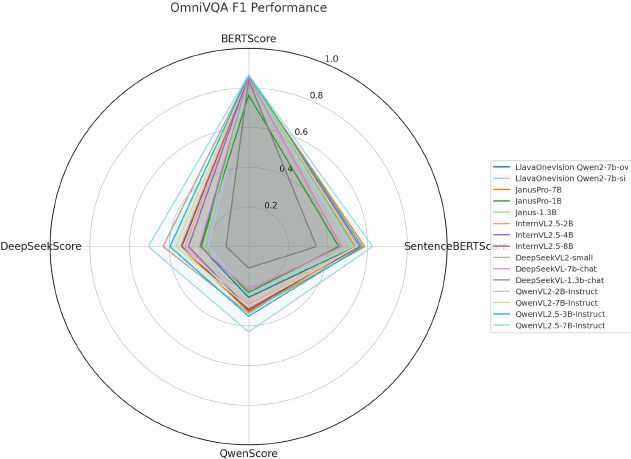}
    \caption{Radar Chart of F1 (Harmonic Mean of Reasoning and Answer) for OmniVQA Benchmark Models.}
    \label{fig:radar-f1}
\end{figure}

\noindent
\textbf{Description:}  
This chart depicts the F1 harmonic mean of reasoning and answer scores for each model across the four metrics. Larger, more outward-stretched polygons correspond to models with better balanced overall performance. QwenVL2.5-7B-Instruct (sky-blue) achieves the outermost values on all axes, confirming it as the top performer across all evaluation metrics. Other models like JanusPro-7B (orange) and InternVL2.5-8B (brown) also show relatively wide coverage, though still lagging behind the QwenVL2.5-7B. In contrast, smaller-capacity models (e.g., JanusPro-1B) have tightly inner-bound polygons, underscoring the significant performance gap between the best model and lower-tier models in the OmniVQA benchmark.

\subsection*{B.4 Reward and Completion Statistics During RL Training}

This section presents the learning curves for key reward signals and completion statistics over the course of RL post-training. Each plot corresponds to a major reward or output property, providing insight into how the model adapts and improves throughout training.

\begin{figure}[H]
    \centering
    \includegraphics[width=0.9\linewidth]{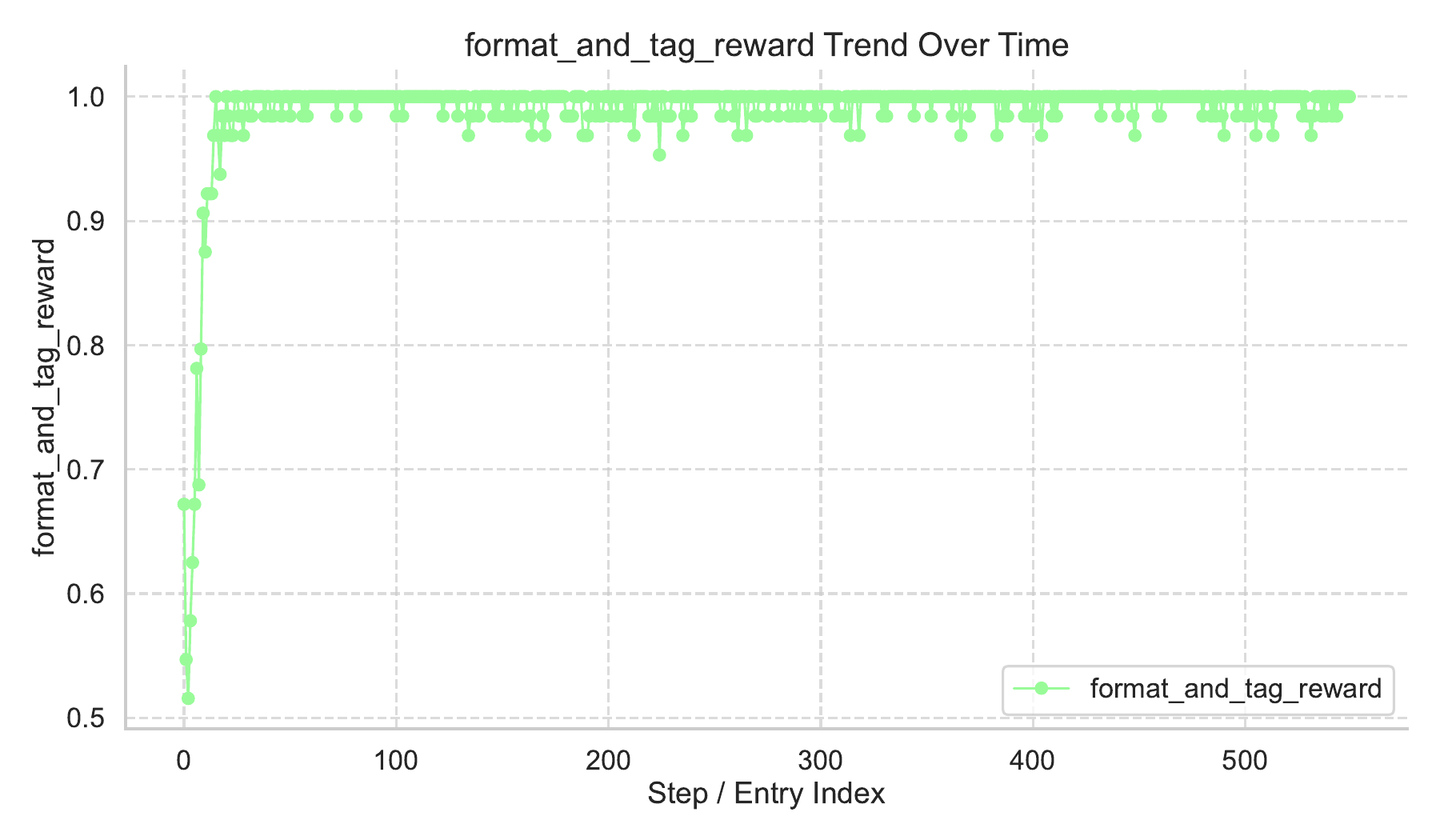}
    \caption{Format and Tag Reward Across Training Steps.}
    \label{fig:format-tag-reward}
\end{figure}
\noindent
\textbf{Interpretation:}
Figure~\ref{fig:format-tag-reward} tracks the \emph{format\_and\_tag\_reward} as a function of training steps (0–550). Initially, the reward hovers near 0.50, indicating that early model outputs only partially comply with the desired formatting and tagging standards. During the first 200 steps, the reward rises quickly above 0.75, then continues a steady upward trend, reaching close to 0.98 by step 500. This near-linear increase demonstrates that the model rapidly learns and maintains strong adherence to prescribed output formatting.

\vspace{2ex}

\begin{figure}[H]
    \centering
    \includegraphics[width=0.9\linewidth]{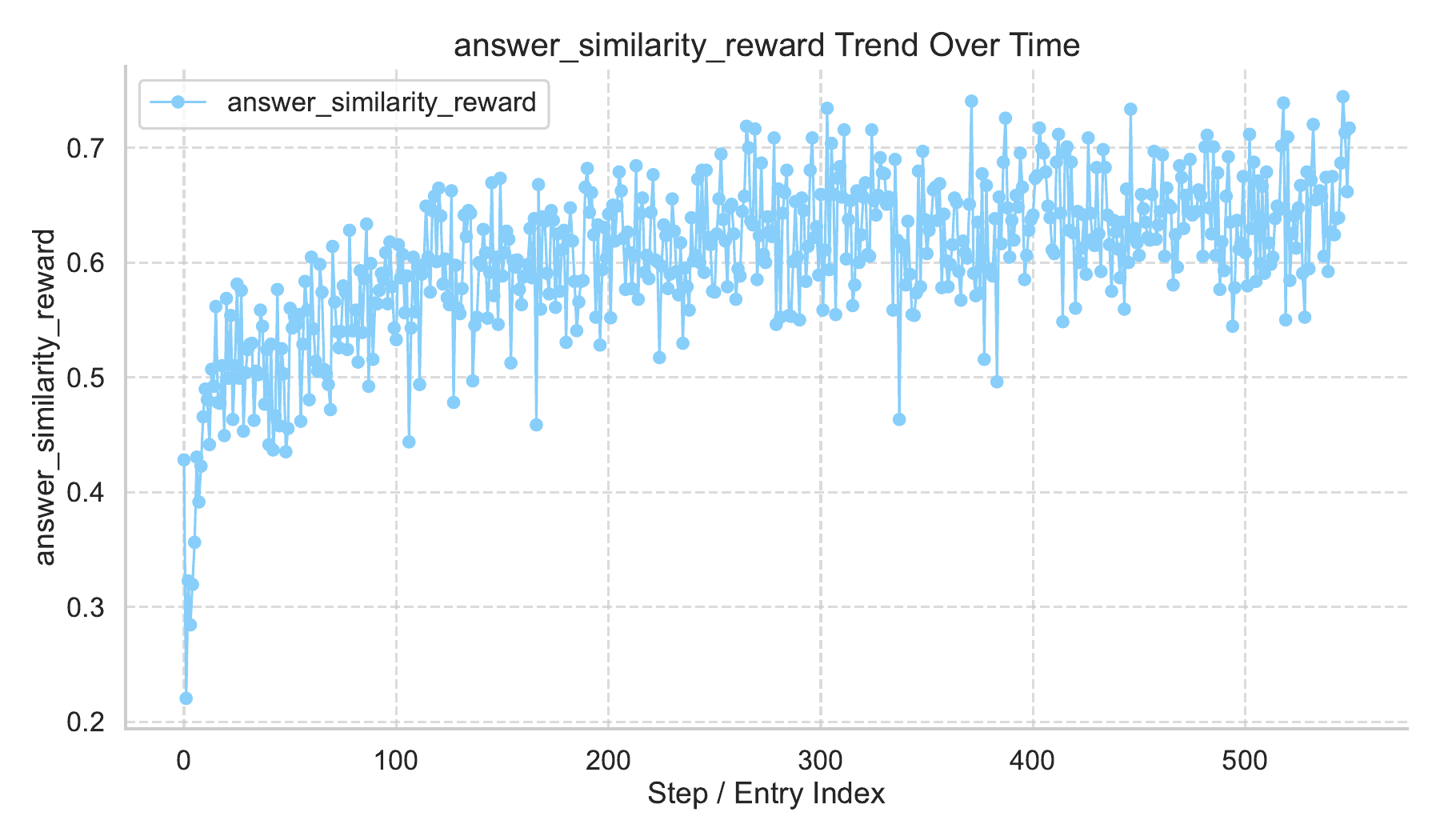}
    \caption{Answer Similarity Reward Over Training Steps.}
    \label{fig:answer-sim-reward}
\end{figure}
\noindent
\textbf{Interpretation:}
Figure~\ref{fig:answer-sim-reward} illustrates the progression of the \emph{answer\_similarity\_reward} during training. The curve starts at approximately 0.25, reflecting a substantial gap between initial model answers and reference responses. Over the first 100–150 steps, the reward increases sharply to about 0.45, then climbs more gradually to around 0.65 by step 500. This consistent upward trend indicates that the model’s generated answers become increasingly semantically and lexically aligned with the ground-truth references as training advances.

\vspace{2ex}

\begin{figure}[H]
    \centering
    \includegraphics[width=0.9\linewidth]{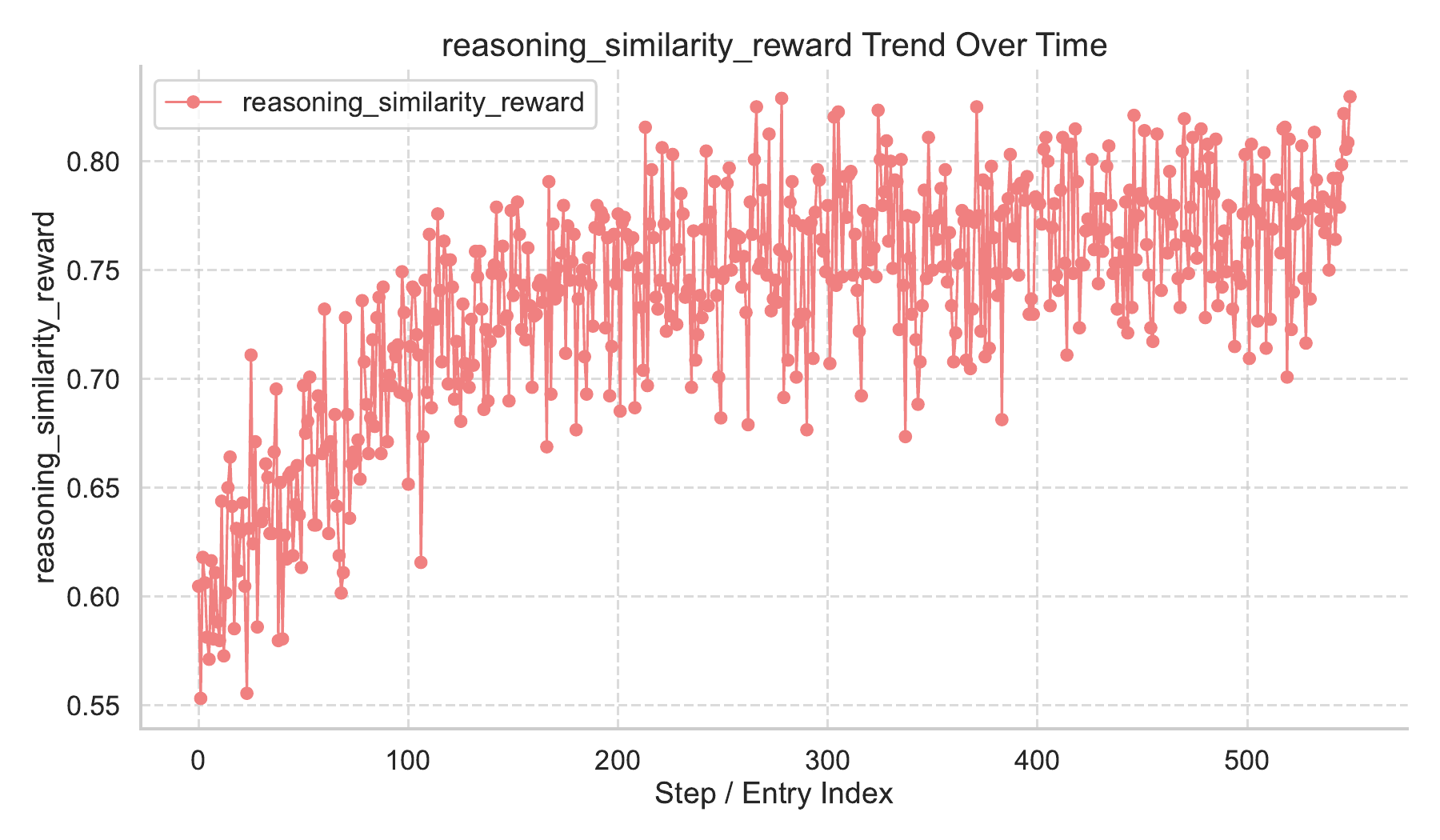}
    \caption{Reasoning Similarity Reward Over Training Steps.}
    \label{fig:reasoning-sim-reward}
\end{figure}
\noindent
\textbf{Interpretation:}
Figure~\ref{fig:reasoning-sim-reward} presents the trajectory of the \emph{reasoning\_similarity\_reward}. Starting at around 0.55—indicating modest alignment between generated and reference reasoning chains—the reward rises most rapidly within the first 150 steps, then continues to grow steadily, approaching 0.80 by step 500. This pattern demonstrates effective learning of structured, expert-like reasoning, with the final high reward signifying close conformity to the reference chains of thought.

\vspace{2ex}

\begin{figure}[H]
    \centering
    \includegraphics[width=0.9\linewidth]{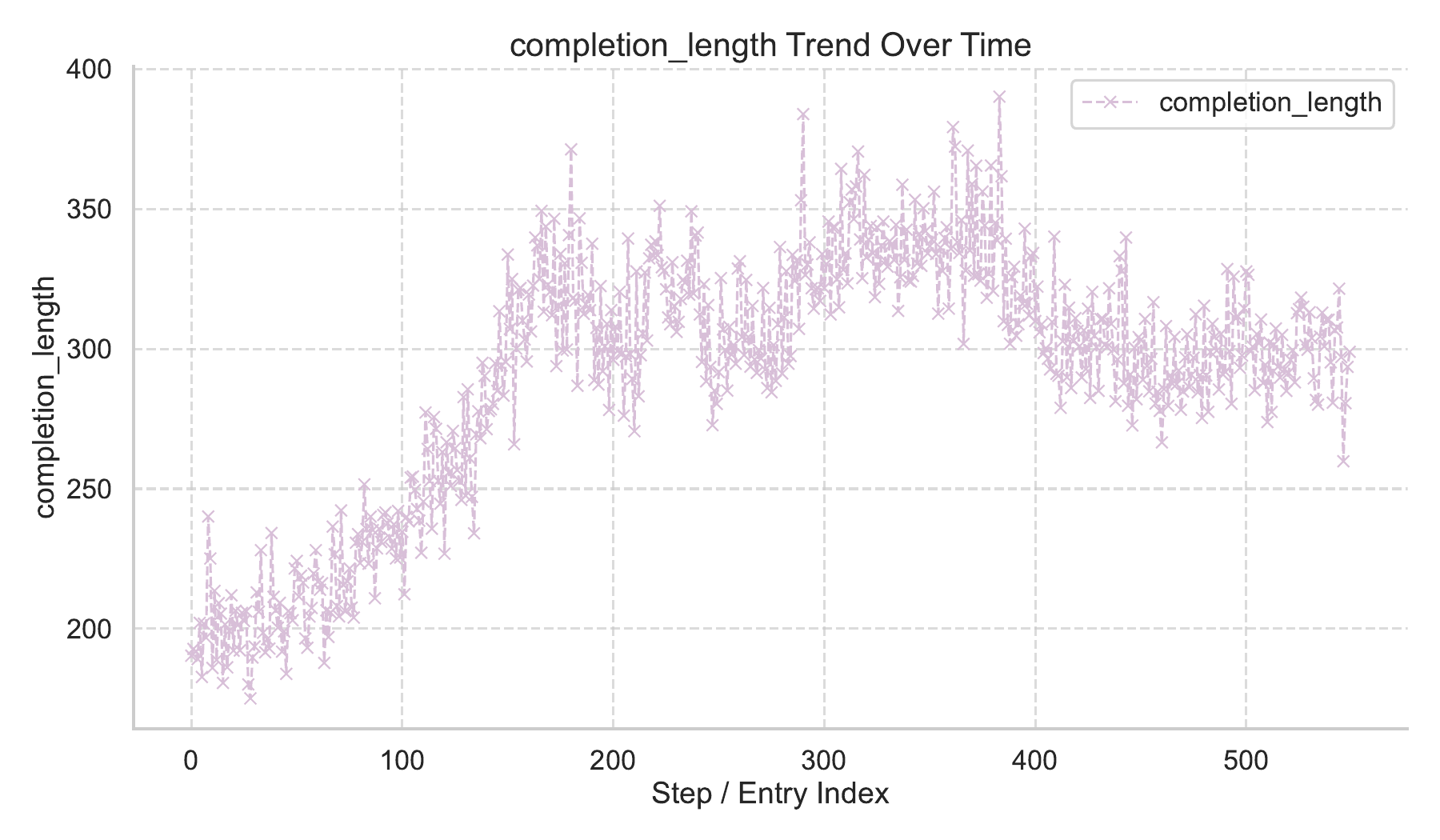}
    \caption{Generated Completion Length Over Training Steps.}
    \label{fig:completion-length}
\end{figure}
\noindent
\textbf{Interpretation:}
Figure~\ref{fig:completion-length} shows the trend in generated completion length (measured in tokens or characters) over training. Early outputs average around 250 units, corresponding to relatively brief responses. Between steps 100 and 300, there is a marked increase to roughly 325, as the model produces more detailed and elaborate outputs. After step 300, length stabilizes in the 350–380 range, reflecting a balance between completeness and conciseness. This upward shift highlights the model’s enhanced capacity to provide comprehensive, well-explained answers.

\end{document}